\documentclass{uai2024} 
                        
\usepackage[american]{babel}

\usepackage{natbib} 
\usepackage{mathtools} 
\usepackage{booktabs} 
\usepackage{tikz} 

\usepackage{tabularx}
\usepackage{amsmath}
\usepackage{amssymb}
\usepackage{amsthm}
\usepackage{graphicx}
\usepackage{subfig}
\usepackage{float}
\usepackage{amssymb,amsfonts}
\usepackage{textcomp}
\usepackage{xspace}
\usepackage{makecell}
\usepackage{booktabs,siunitx}
\usepackage{diagbox}
\usepackage{array, multirow}
\usepackage{newfloat}
\usepackage{listings}
\usepackage{caption}
\usepackage{times}
\usepackage{soul}
\usepackage{url}

\title{Intriguing 
Differences Between Zero-Shot and Systematic Evaluations 
of Vision-Language Transformer Models
}

\author[1]{Shaeke Salman\thanks{Equal contributions}} 
\author[1]{Md Montasir Bin Shams\textsuperscript{*}}
\author[1]{Xiuwen Liu}
\author[2]{Lingjiong Zhu}
\affil[1]{%
    Department of Computer Science\\
    Florida State University\\
    Tallahassee, Florida, USA 
}
\affil[2]{%
    Department of Mathematics\\
    Florida State University\\
    Tallahassee, Florida, USA  
}

\affil[e]{%
    \{salman, liux\}@cs.fsu.edu, mshams@fsu.edu, zhu@math.fsu.edu
  }
\begin{document}
\maketitle

\begin{abstract}

Transformer-based models have dominated natural language processing and other areas in the last few years due to their superior (zero-shot) performance on benchmark datasets. However, these models are poorly understood due to their complexity and size. While probing-based methods are widely used to understand specific properties, the structures of the representation space are not systematically characterized; consequently, it is unclear how such models generalize and overgeneralize to new inputs beyond datasets. In this paper, based on a new gradient descent optimization method, we are able to explore the embedding space of a commonly used vision-language model. Using the Imagenette dataset, we show that while the model achieves over 99\% zero-shot classification performance, it fails systematic evaluations completely.  
Using a linear approximation, we provide a framework to explain the striking differences. We have also obtained similar results using a different model to support that our results are applicable to other transformer models with continuous inputs. We also propose a robust way to detect the modified images.  

\end{abstract}

\section{Introduction}

Transformers have emerged as a dominant approach in various natural language processing tasks in recent years, resulting in significant performance enhancements~\citep{vaswani2023attention,Devlin2018Bert,radford2018improving,colin2020t5}. Their remarkable capabilities have also led to widespread adaption in the field of computer vision and other domains~\citep{dosovitskiy2021image,henaff2020data,mainu2023detect,emdad2023towards,biswas2022geometric,li2023graph,baevski2020wav2vec}. Consequently, the recent surge in multimodal models aims to leverage the exceptional performance of transformer models across various modalities, including visual, textual, and more~\citep{xu2023multimodal,zhu2023minigpt,openai2023gpt4,girdhar2023imagebind}. 
Despite their proficiency in zero-shot settings across various applications, how such models generalize systematically remains unknown.

\begin{figure*}[ht]
  \centering
  \includegraphics[width=0.90\textwidth]  {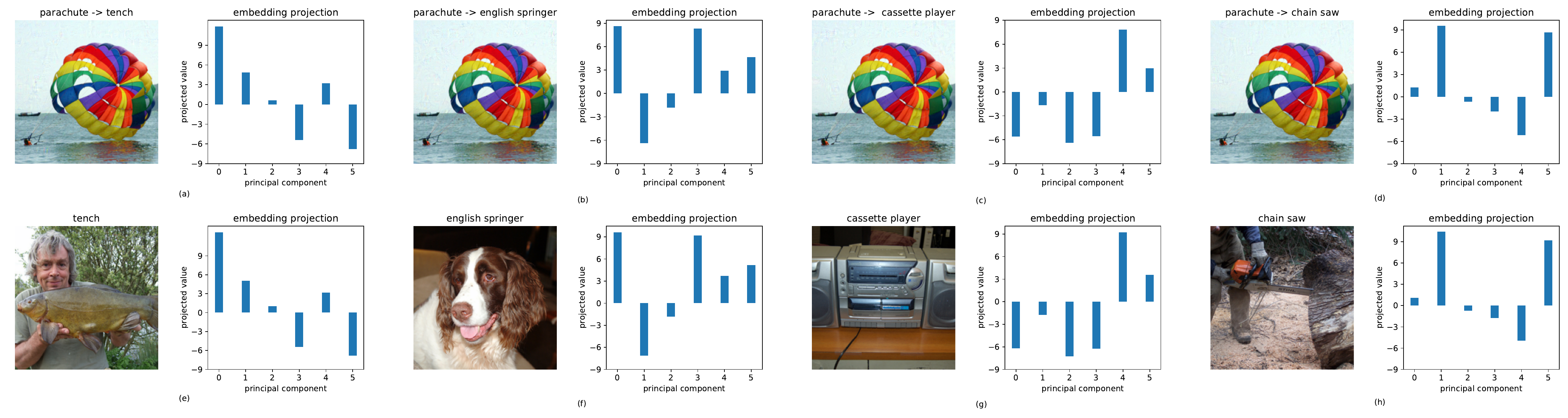}\label{fig:projection_embeddings}
  
  \includegraphics[width=0.90\textwidth]{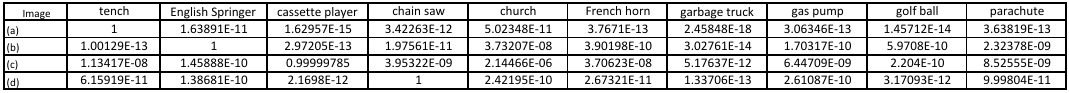}\label{fig:vt_mat_parachute}
  \caption{Typical examples from ImageNet obtained using the proposed framework. The visually indistinguishable images have different representations from each other as shown in their low-dimensional projections.  Note that the arrow in the title ($original \rightarrow target$) signifies a derived image from the original one by aligning the embedding of the original image with the target image using our method. The projections of embedding-aligned images closely resemble the projections of the aligned class. The matrix shows the classification outcomes from the multimodal ImageBind pretrained model used directly with no modifications; Please refer to the supplementary materials for all the nine embedding-aligned images, projections, and the full $vision \times text$ matrix.
  }
  \label{fig:overall}
\end{figure*}

As multimodal models allow continuous inputs, they also make gradient descent and certain mathematical analyses (such as linear approximations) applicable. In this paper, through a gradient-descent-based optimization procedure, we demonstrate that, for each image, there exist subspaces comprising images that are practically indistinguishable from the original despite their substantially different representations. Fig.~\ref{fig:overall} demonstrates such cases, where four visually indistinguishable parachute images in Fig.~\ref{fig:overall}, (a), (b), (c), and (d), have very different representations as shown by their low-dimensional projections. These indistinguishable images are generated by aligning the representation of an image originally belonging to the parachute class with the representation from other classes, as depicted in Fig.~\ref{fig:overall} (e), (f), (g), and (h), respectively. One can see the projections of the embedding-aligned parachute images have a striking resemblance to the projections of the aligned class, despite their substantial semantic differences.

When we pass these images to an unmodified multimodal ImageBind model\footnotemark\footnotetext{https://github.com/facebookresearch/ImageBind.}~\citep{girdhar2023imagebind}, visually indistinguishable images are confidently classified into every other class within the dataset as shown in the vision $\times$ text matrix in Fig.~\ref{fig:overall}; for a comprehensive view of the nine embedding-aligned images for all the classes, their projections, and the complete vision $\times$ text matrix, please refer to the supplementary materials. This observation holds true for all images in the imagenette dataset\footnotemark\footnotetext{Obtained from https://github.com/fastai/imagenette.}. Therefore, for each image, we can find subspaces consisting of images embedding-aligned with other classes and visually indistinguishable; simultaneously, their representations match that of other classes. Thus, while an ImageBind model may perform reliably in a zero-shot setting, it struggles to generalize robustly. This paper offers a clear demonstration of this on the imagenette dataset, with implications that extend to similar datasets. Since our results do not depend on model-specific details, they generalize to other transformer-based models.

Our main contributions are as follows:
\begin{itemize}
  \item Using the proposed efficient computational procedure to match specified representations, we clearly demonstrate that while transformer-based vision-language models give very high zero-shot accuracy on a diverse set of images, close to every image in the dataset, there are visually indistinguishable images
  that are classified to every other class in the dataset with high confidence. 
  \item Using a linear approximation of the representation function locally, we show that adding Gaussian noise in the image space leads to a normal distribution for pairwise classification and therefore the vision-language models may not generalize robustly. 
  \item Our results show that zero-shot learning performance does not imply robust generalization and the models must be evaluated systematically. 
\end{itemize}

\section{Related Work}

Recent advancements in the field of natural language processing have inspired the development of a novel deep learning approach known as Vision-Language Model (VLM) pretraining and zero-shot prediction, which has gained significant attention~\citep{radford2021learning}. Notably, it introduces the concept of zero-shot prediction, suggesting that the pretrained VLMs can make predictions for tasks it has never encountered during training. For example, the recently proposed CLIP model~\citep{radford2021learning} utilizes image-text contrastive objectives to achieve such capabilities. While these models tend to work well in zero-shot settings, their generalization to unseen inputs is not well understood. Understanding how a model generalizes and potentially overgeneralizes is crucial before considering their widespread deployment in real-world applications. To comprehend this, it is important to identify which inputs share the same representations, as downstream applications will treat them the same. In our study, we demonstrate that transformer-based vision-language models exhibit high zero-shot classification performance across a diverse image set, yet they encounter difficulties in classifying visually similar images, highlighting the challenges associated with consistent generalization.

There has been a notable trend in the use of large multimodal models, which offer significant advantages by employing a unified embedding space for various data modalities. Some of the state-of-the-art multimodal models include GPT-4~\citep{openai2023gpt4} , MiniGPT-4~\citep{zhu2023minigpt}, Flamingo~\citep{alayrac2022flamingo}, Bard~\citep{Bard}, LLaVA~\citep{llava}, and ImageBind~\citep{girdhar2023imagebind}. Despite recent studies that have improved the performance of these transformed-based models on standard datasets and various tasks, the fundamental problems of understanding how these models generalize, overgeneralize, and memorize information continue to pose an ongoing challenge~\citep{Zhang2016UnderstandingDL,zhang2017understanding,Neyshabur2017ExploringGI}. Given the widespread use of these multimodal models, a deeper understanding of how they generalize becomes increasingly significant. Although these models have architectural differences, they all share the common foundation of being transformer models. Consequently, the framework and findings outlined in our paper also directly apply to these models. Our findings demonstrate that such models exhibit impressive performance on diverse tasks (e.g., zero-shot classification), however, their inherent generalization abilities are limited by the properties of the underlying embedding spaces. 

Another line of research aimed at gaining a deeper understanding of the transformer-based models involves probing them to discover new characteristics. Probing techniques have been widely used in the context of understanding transformer models when applied to discrete inputs, such as text~\citep{niven-kao-2019-probing,clark-etal-2019-bert}. 
Adversarial attacks are closely related, where unnoticeable changes to the input can cause the models to change their predictions. Neural networks are well-known targets for adversarial attacks, where most of these attacks exploit particular characteristics of the classifier. 
Recent studies have addressed adversarial attacks on multimodal models, which can potentially jailbreak aligned LLMs or VLMs~\citep{carlini2023aligned,qi2023visual,zou2023universal}. 
Several recent works have focused on evaluating the robustness of Vision Transformers (ViTs) against attacks that involve accessing the internal architecture of the models~\citep{bhojanapalli2021understanding,qin2023understanding,herrmann2022pyramid}.
Furthermore, in the case of computational pathology,  ~\citet{Laleh2022.03.15.484515} note that both the baseline and adversarially trained variants of ViTs outperform Convolutional Neural Networks (CNNs) in terms of robustness. Notably, these studies involve generating adversarial examples based on the classifier's methodology rather than focusing on representations. While our method can indeed generate effective adversarial examples, our approach differs significantly. We do not create an adversarial example for a specific classifier. Instead, our method can generate examples that align with a specified representation, and a subset of such inputs shares the same property. In this paper, our primary focus is to study the semantics of local representations provided by vision-language models, not specific adversarial attacks. We accomplish this by leveraging the space by gradient descent optimization and conducting in-depth mathematical analyses.

\section{Preliminaries}
\label{sec:preli}
As this paper focuses on vision-language models based on transformers, here we first describe the transformers mathematically and then describe the vision-language models. Transformers can be described mathematically succinctly, consisting of a stack of transformer blocks.
A {transformer block} is a parameterized function class $f_\theta: \mathbb{R}^{n \times d} \rightarrow \mathbb{R}^{n \times d}$. If $\mathbf{x} \in \mathbb{R}^{n \times d}$ then $f_\theta(\mathbf{x}) = \mathbf{z}$ where
$Q^{\left(h\right)}\left(\mathbf{x_i}\right) = W^T_{h,q}\mathbf{x}_i,\quad
    K^{\left(h\right)}\left(\mathbf{x_i}\right) = W^T_{h,k}\mathbf{x}_i,\quad
    V^{\left(h\right)}\left(\mathbf{x_i}\right) = W^T_{h,v}\mathbf{x}_i,\quad
    W_{h,q}, W_{h,k}, W_{h,v} \in \mathbb{R}^{d \times k}$.
The key multi-head self-attention is a softmax function applying row-wise on the inner products.\footnote{Note that there are other ways to compute the attention weights.}
\begin{equation}
    \alpha_{i,j}^{\left(h\right)} = \texttt{softmax}_j\left(\frac{\left<Q^{\left(h\right)}\left(\mathbf{x}_i\right),K^{\left(h\right)}\left(\mathbf{x}_j\right)\right>}{\sqrt{k}}\right).
\end{equation}
The outputs from the softmax are used as weights to compute new features, emphasizing the ones with higher weights given by
\begin{equation}
    \mathbf{u}'_i = \sum\limits_{h=1}^H W^T_{c,h} \sum\limits_{j=1}^n \alpha_{i,j} V^{\left(h\right)} \left(\mathbf{x}_j\right),\quad
    W_{c,h} \in \mathbb{R}^{k \times d}.
\end{equation}

The new features then pass through a layer normalization, followed by a ReLU layer, and then another layer normalization. 
Typically transformer layers are stacked to form deep models. 

While transformer models are widely used for natural language processing tasks, recently, they are adapted to vision tasks by using image blocks on the basic units, and spatial relationships among the units are captured via the self-attention mechanism~\citep{dosovitskiy2021image}. A vision-language model based on transformers incorporates a dedicated transformer model for each input modality. The resulting representations from these modalities are mapped to a shared embedding space.

For the ImageBind model, we denote the model for image $x$ as $f(x)$ and for text $t$ as $f_t(t)$. Given image $x$ and $C$ text labels, $t_0, \ldots, t_{C-1}$, the zero-shot classification uses softmax applied on the dot products of the image and text representations. Therefore, the probability classified $x$ to $t_i$ is given by
\begin{equation}
\frac{e^{f(x)^T f_t(t_i)}}{\sum_{j=0}^{C-1}e^{f(x)^T f_t(t_j)}}.
\end{equation}
Note the probabilities reported in the figures in this paper are computed using a publicly available ImageBind model without any change. 
While the proposed method applies to all transformer-based models with continuous inputs, we focus on the CLIP model~\citep{radford2021learning}, which jointly models images and text using the same shared embedding space used in the ImageBind model.

\section{Methodology}
As introduced in our earlier work~\citep{salman2024intriguing}, we have developed a framework that allows us to explore the embedding space, analyze their properties, and verify them in large models. 
In general, we model the representation given by a (deep) neural network (including a transformer) as a function $f: \mathbb{R}^m\rightarrow\mathbb{R}^n$. A fundamental question is to have a computationally efficient and effective way to explore the embeddings of inputs by finding the inputs whose representation will match the one given by $f(x_{tg})$, where $x_{tg}$ is an input whose embedding we like to match. Informally, given an image of a parachute in Fig.~\ref{fig:overall} as an example, all the images that share its representation given by a model will be treated as a parachute. Furthermore, understanding the structures of the representation space is crucial, as they determine how the model generalizes. For a thorough understanding of the framework, we recommend referring to our earlier research as detailed in ~\citet{salman2024intriguing}. 

\subsection{Embedding Matching Procedure}
As described in our earlier work~\citep{salman2024intriguing}, the proposed approach for embedding alignment aims at matching the representation of an input to that of a target input. 
To accommodate the requirement of aligning two vectors, we define the loss for finding an input matching a given representation as
\begin{equation}
   L(x) = L(x_0+\Delta x)= \frac{1}{2}\Vert f(x_0+\Delta x) - f(x_{tg})\Vert^2,
\end{equation}
where $x_0$ represents the initial input and $f(x_{tg})$ defines the target embedding. The gradient is approximately determined by
\begin{equation}
\frac{\partial L}{\partial x} \approx \left(\frac{\partial f}{\partial x}\Big|_{x=x_0}\right)^T(f(x_0+\Delta x) - f(x_{tg})).
\label{eq:grad_J}
\end{equation}
Eq.~\eqref{eq:grad_J} demonstrates the relationship between the gradient of the mean square loss function and the Jacobian of the representation function evaluated at $x=x_0$. Although obtaining optimal solutions might involve solving either a quadratic or linear programming problem, depending on the chosen norm for minimizing $\Delta x$, the gradient function has proven to work effectively for all the cases we have tested, attributable to the Jacobian of the transformer.

A practical challenge in employing the gradient descent-based method is selecting an appropriate learning rate. For transformers, the model can be approximated by a linear model when it moves within one activation region, despite inherent nonlinearities introduced by softmax and ReLU functions. This approximation allows the algorithm to find matching representations across a wide range of learning rates, as detailed further in Section 5.


\subsection{The Effects on Normal Distributions}
\label{sub:normal}
To understand how a transformer-based model generalizes, one needs to characterize how a probability distribution in the input space will be transformed into the representation space. 
Due to the complexity and size of the transformer-based models, answering the question is challenging. As the first attempt to tackle the issue, here we like to understand how adding Gaussian noise to an input image will affect the representation and the classification. 
Given an input $x_0$, the local structures decide how the model behaves in the local neighborhood; 
the linear approximation of the function is given by
\begin{equation}
    f(x_0+\Delta x) \approx f(x_0) + \frac{\partial f}{\partial x}\Big|_{x=x_0} \times \Delta x.
\end{equation}
As in Eq.~\eqref{eq:grad_J}, $\frac{\partial f}{\partial x}$ is the Jacobian matrix of the function at $x=x_0$.
As a result, for deployed models, where $m>n$, there is a null space where the embeddings do not change as the input changes; it can be obtained via a reduced singular decomposition of the Jacobian. There is a normal space in the space perpendicular to the null space, where the embeddings can change quickly. 
In the case of adding i.i.d. Gaussian noise with standard deviation $\sigma_n$ to every pixel, $\Delta x$ will be a normal distribution as well. The covariance matrix of the pixels in the noise-added image would be $\sigma_n \times I$, where $I$ is the identity matrix. Under the linear assumption, the resulting probability distribution in the representation space will be normal as well, whose mean is $f(x_0)$, and whose variance matrix is given by
\begin{equation}
\left(\frac{\partial f}{\partial x}\right)\left(\frac{\partial f}{\partial x}\right)^T\sigma_n^2
= M_J \times \sigma_n^2.
\label{eq:var_trans}
\end{equation}

To be able to model how the noisy images will be classified by the ImageBind model, we consider two classes $t_0$ and $t_1$. As the classification is done by applying the softmax on the inner dots between the representation of image $x$ and the embeddings for $t_0$ and $t_1$, denoted as 
$f_t(t_0)$ and $f_t(t_1)$. The classification probability that $x$ belongs to $t_1$ is given by 
\begin{equation}
\frac{e^{f(x)^T f_t(t_0)}}{e^{f(x)^T f_t(t_0)}+e^{f(x)^T f_t(t_1)}} =
\frac{e^{f(x)^T (f_t(t_0)-f_t(t_1))}}{e^{f(x)^T (f_t(t_0)-f_t(t_1))}+1}.
\label{eq:two_classes}
\end{equation}

Therefore the binary classification is determined by whether $f(x)^T (f_t(t_0)-f_t(t_1)) > 0 $ or not. Now if we generate a sample by adding Gaussian noise to $x$, assuming the local linear approximation of the transformer model for $x$, the resulting probability distribution would be a normal
distribution with $f(x)^T (f_t(t_0)-f_t(t_1))$ being the mean and $(f_t(t_0)-f_t(t_1))^T\times M_J\times
(f_t(t_0)-f_t(t_1)) \times\sigma_n^2$ as the variance.
With this, we can predict the accuracy of classifying
these noisy samples. If the mean is away from zero
and the variance is relatively small, the added noise will not affect the classification. Increasing the variance of Gaussian noise will decrease the classification accuracy. This is consistent with the empirical results using examples from the Imagenette
dataset shown in the next section.

\begin{figure*}[ht]
  \centering
  \includegraphics[width=0.80\textwidth]{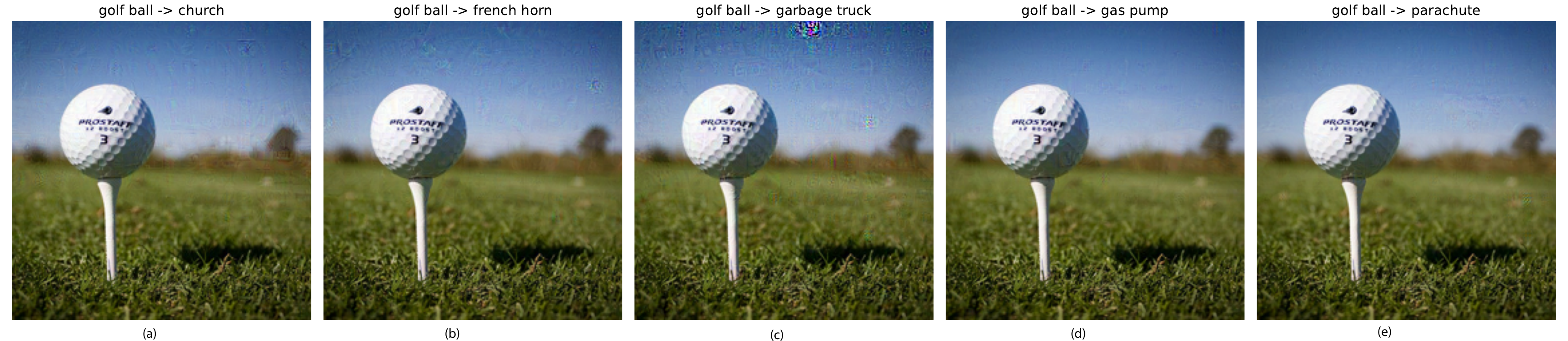}\label{fig:more_matched}
  \includegraphics[width=0.80\textwidth]{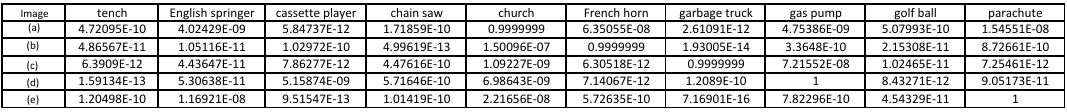}\label{fig:vt_mat_golf}
  
  \caption{
  More examples where visually indistinguishable images have very different representations via embedding alignment and therefore very different classification outcomes as shown in the classification probabilities.
}
  \label{fig:overall_golf}
\end{figure*}

\begin{figure}[ht]
  \centering
  \includegraphics[width=0.95\columnwidth,height=1.2in]
  {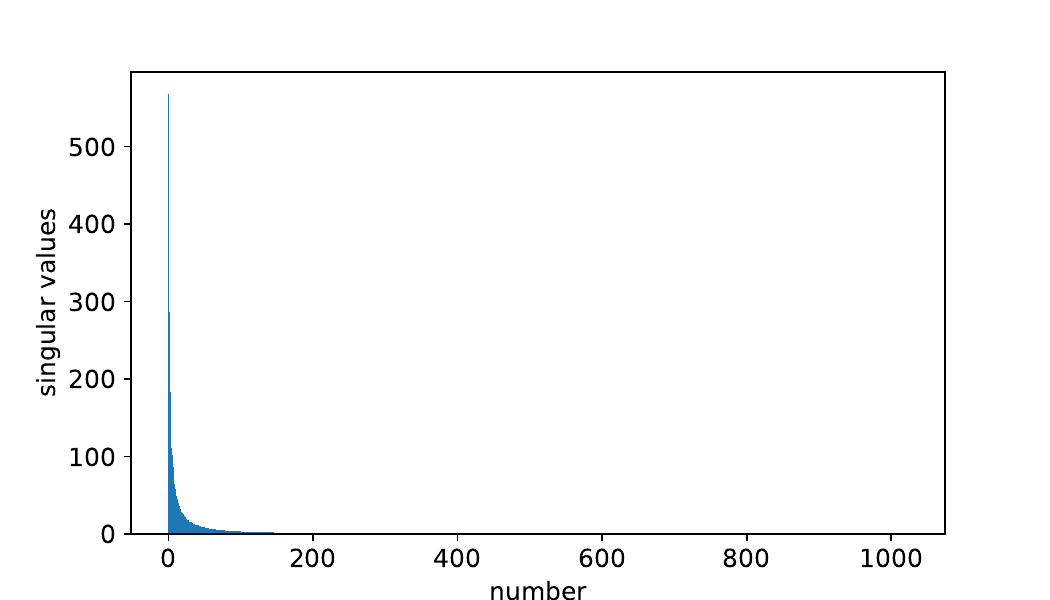} \\ 
  \includegraphics[width=0.95\columnwidth,height=1.2in]
  {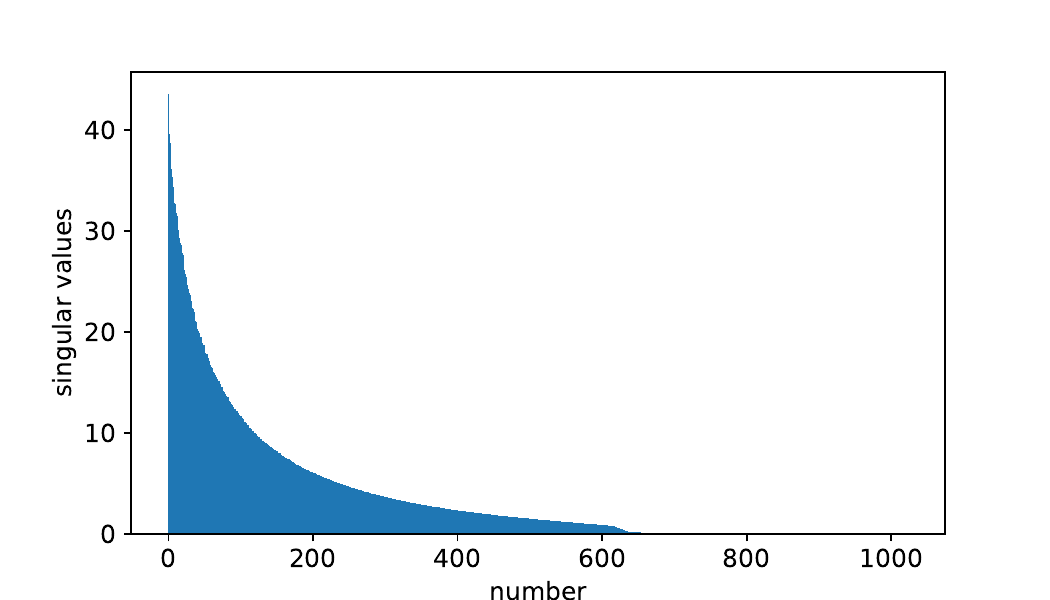} 

  \caption{The singular values of the Jacobian matrix for the leftmost image in Fig.~\ref{fig:overall}(a) (bottom) and that for the original version of the image (top).
  }
  \label{fig:singulars_lc}
\end{figure}

\section{Experiments}
In this section, we initially provide details about the experimental settings and implementation specifics. Our proposed framework is systematically employed on diverse datasets and multiple multimodal models. In the following subsections, we discuss both the outcomes of the experiments and present quantitative results.

\subsection{Datasets and Settings}

\textbf{Datasets:} We perform comprehensive experiments to assess our proposed method using Imagenette, a subset of the ImageNet dataset~\citep{imagenet2009Deng}. This subset comprises images from ten classes of the larger ImageNet dataset. We also show our evaluation with other popular vision datasets, such as MS-COCO~\citep{lin2015microsoft} and Google Open Images~\citep{OpenImages}. 

\textbf{Implementation Details:} To demonstrate the feasibility of the proposed method on large models, we have used the pre-trained model publicly available by ImageBind\footnotemark\footnotetext{https://github.com/facebookresearch/ImageBind}, which in turn uses a CLIP model\footnotemark\footnotetext{https://github.com/mlfoundations/open\_clip}. More specifically, ImageBind utilizes the pre-trained vision (ViT-H 630M params) and text encoders (302M params) from the OpenCLIP~\citep{ilharco_gabriel_2021_5143773,girdhar2023imagebind}. We conduct all experiments on a workstation in our lab equipped with two NVIDIA A5000 GPUs. We will provide source code for all our experiments in GitHub\footnotemark\footnotetext{https://github.com/programminglove08/SystematicEval}.  

The input size is $224\times224\times3$, and the dimension of the embedding is 1024;
for larger images in the dataset, they are cropped to the input size centered at the center of the input ones.
As a result, the Jacobian matrix is of size $1,024\times 150,528$.
Then we use reduced singular value decomposition to write the Jacobian as $U\Sigma V^T=\sum_{i=0}^{1023}s_i\times U(:, i)\times V(:, i)^T$,
where $T$ denotes the matrix transpose operator.
The bottom plot in Fig.~\ref{fig:singulars_lc} shows the singular values of the Jacobian matrix in Fig.~\ref{fig:overall}(a), whereas the top one shows that for the original image. The distribution of the singular values shows that the Jacobians have several dominating directions, reflecting the training set and the training algorithm being used. It is clear that the singular value distributions are significantly different in the original image and the tench-matched one due to the gradient descent optimization in the matching procedure. While we have observed similar patterns on the images we have used, as a future work we are exploring the patterns on the entire ImageNet dataset to understand the underlying mechanism.

\subsection{Experimental Results}

\begin{figure}[ht]
  \centering
  \includegraphics[width=0.9\columnwidth]
  {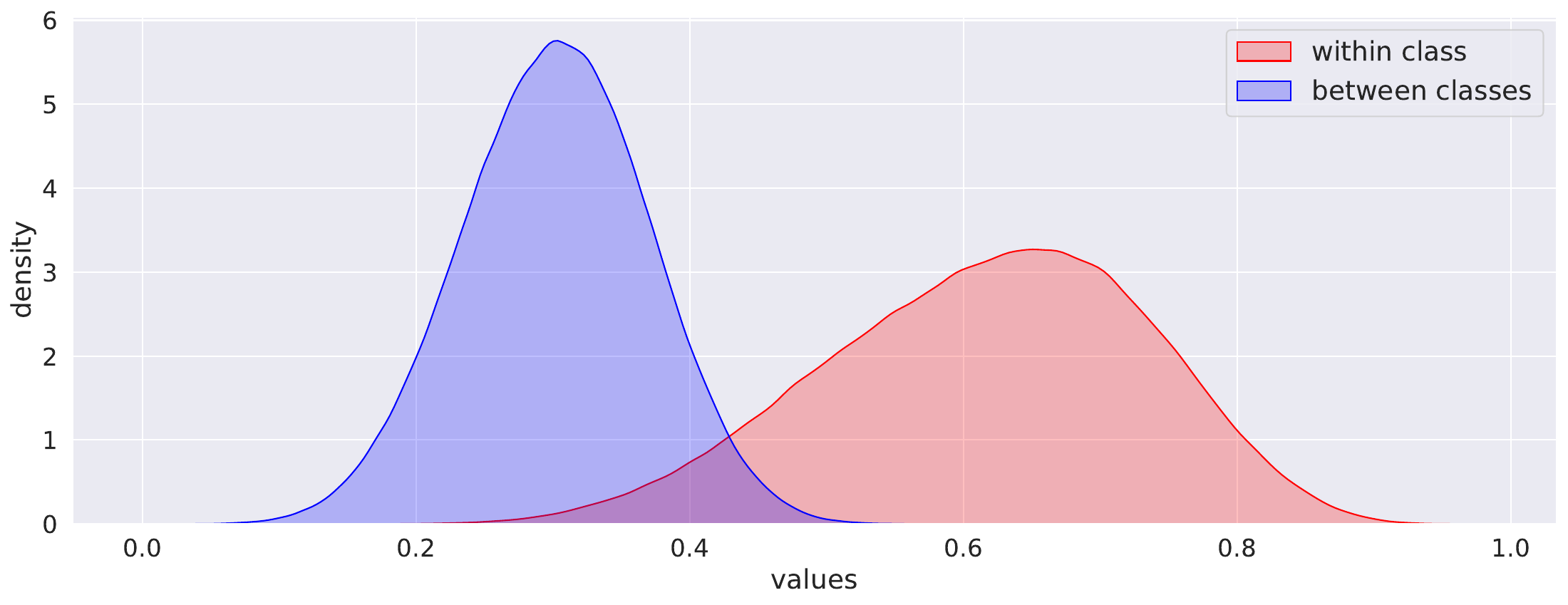} \\ 
  \caption{Comparison of pairwise cosine similarity distribution with the pairs from the same class (red) to that with the pairs from two different classes (purple) from the Imagenette dataset.
}
  \label{fig:similarity_dist}
\end{figure}

\begin{figure}[ht]
  \centering
  \vspace{-0.20in}
  \includegraphics[width=0.90\columnwidth]{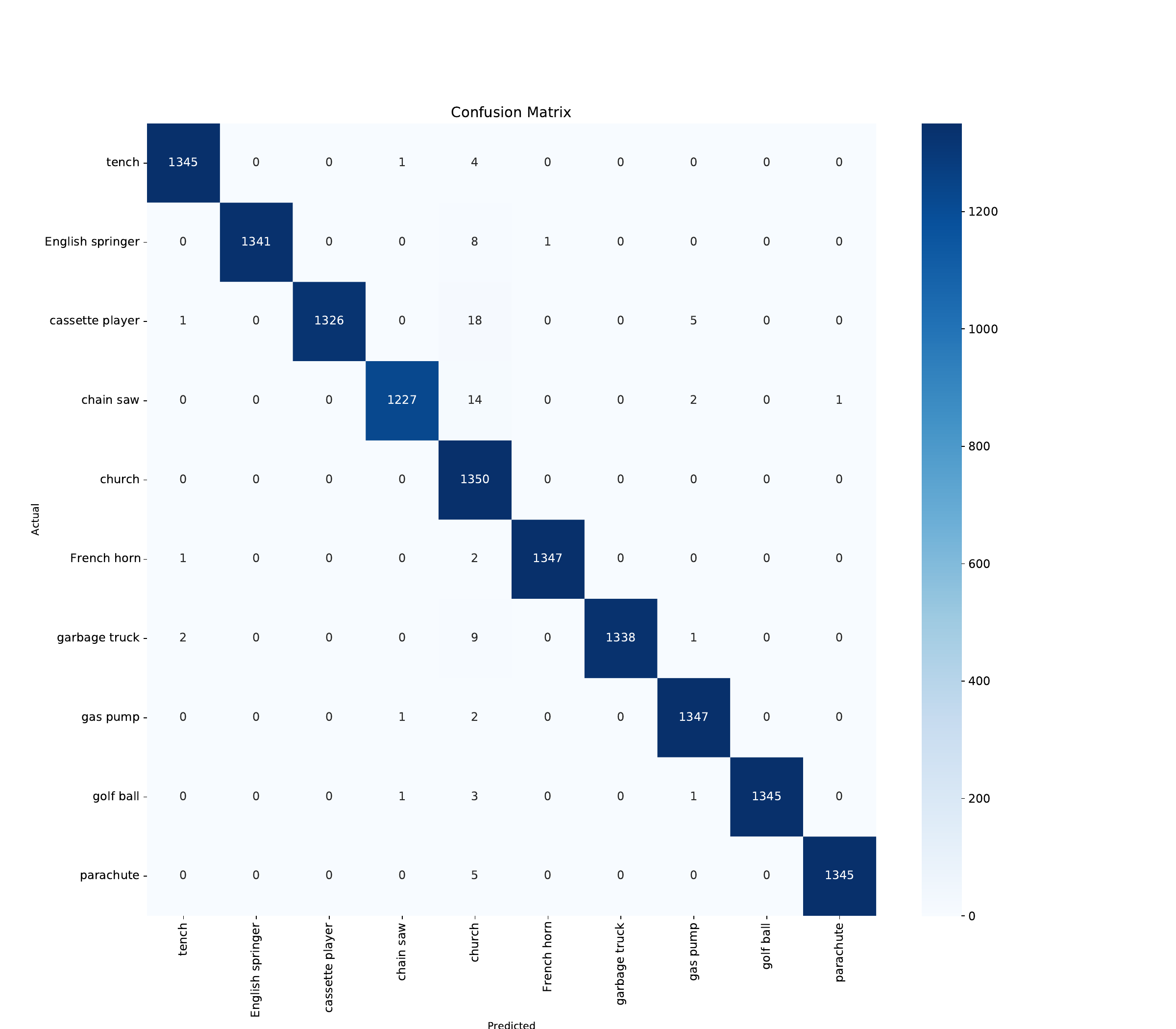}

  \caption{Confusion matrix of zero-shot classification performance for all the images from all the ten classes from Imagenette. The overall accuracy is $99.38\%$ for all the images. 
  }
  \label{fig:conf_matrix}
\end{figure}

\begin{figure}[ht]
  \centering
  \includegraphics[width=0.220\textwidth]{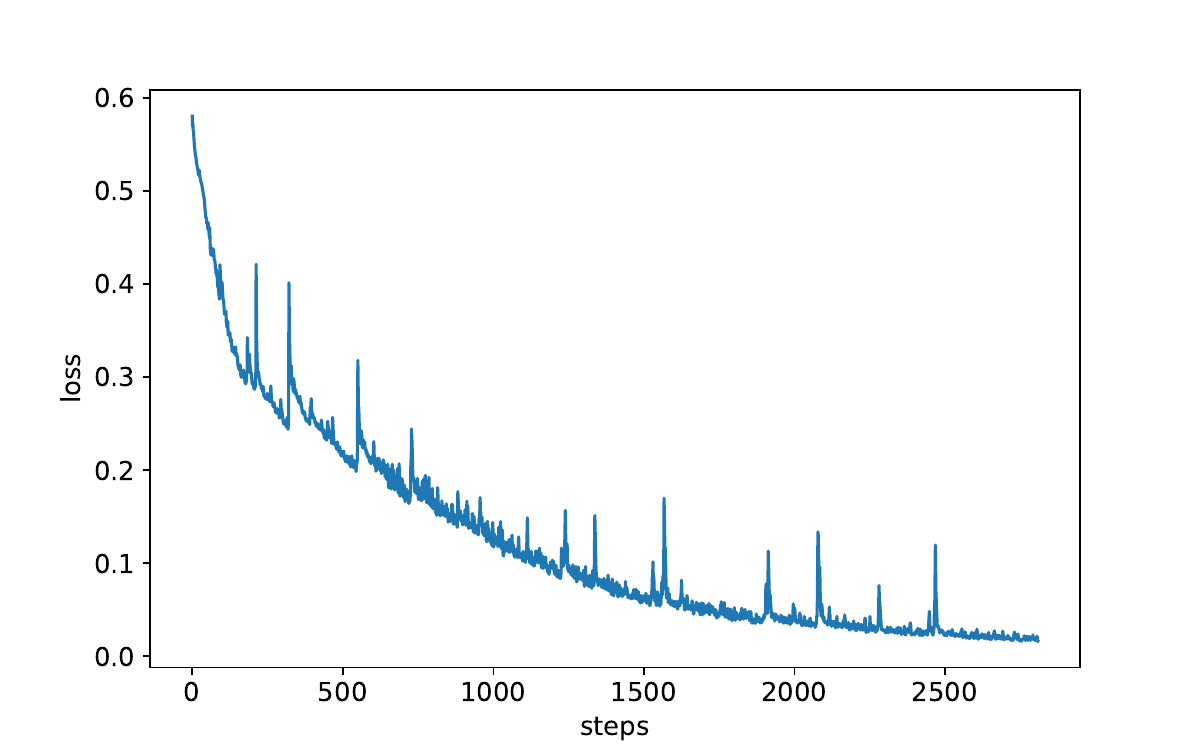}
  \includegraphics[width=0.240\textwidth]{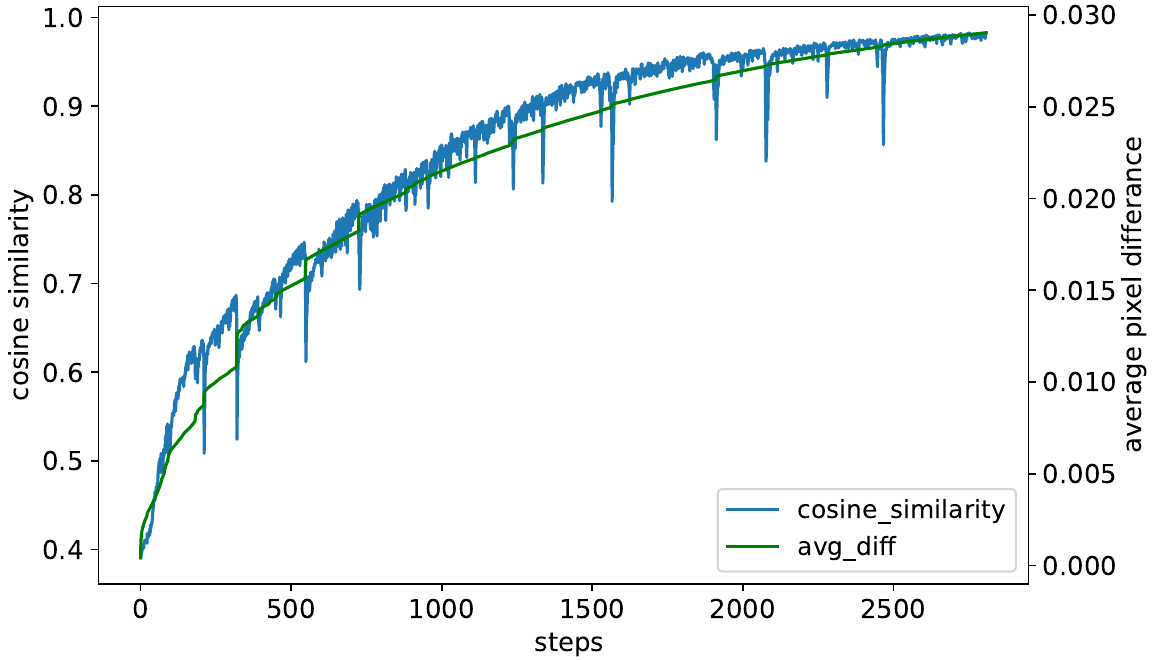}
  
  \caption{The evolution of loss while matching a target embedding. (left) the loss w.r.t. steps. (right) the cosine similarity between the embeddings of the new input and the target w.r.t. the steps, along with the average pixel value difference between the new input and the original image.}
  \label{fig:training_dynamics}
\end{figure}

\begin{figure}[ht]
  \centering
  \includegraphics[width=0.8\columnwidth]{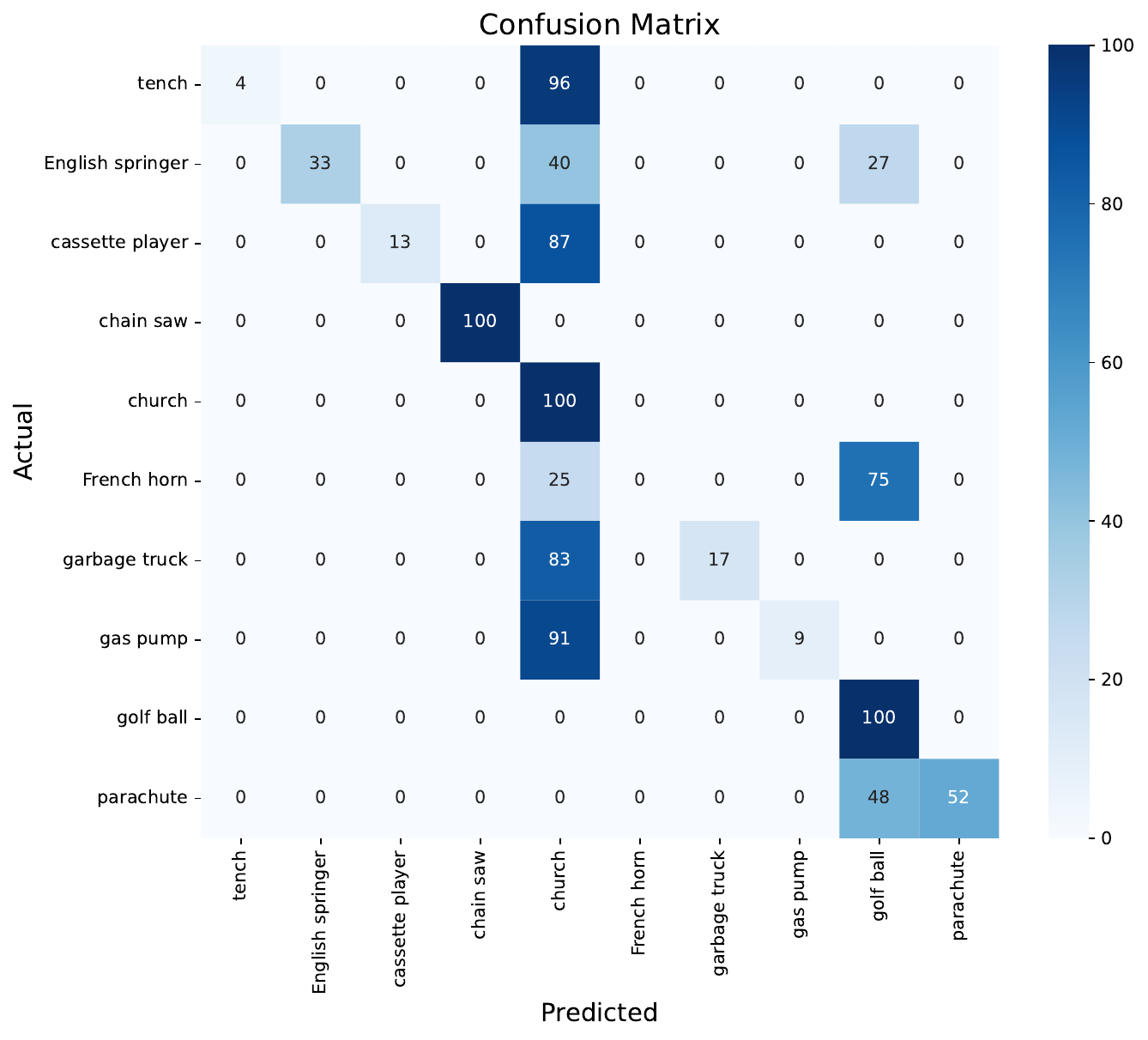}

  \caption{Confusion matrix for 100 noisy images with standard deviation 
  0.50; 
  for each row, 100 noisy images were generated by adding the 100 noise matrices to a single selected image from the specified class.}
  \label{fig:conf_matrix_noisy}
\end{figure}

\begin{figure}[ht]
  \centering
  \includegraphics[width=0.12\textwidth]
  {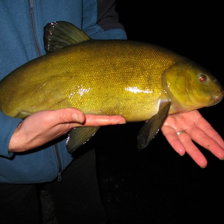}
  \includegraphics[width=0.12\textwidth]
  {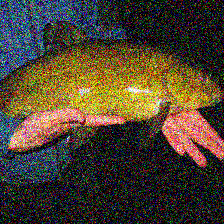}
  \includegraphics[width=0.12\textwidth]
  {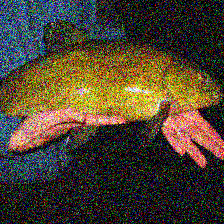}\\

  \includegraphics[width=0.12\textwidth]
  {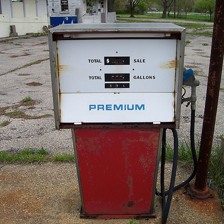}
  \includegraphics[width=0.12\textwidth]
  {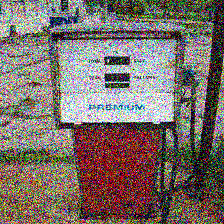}
  \includegraphics[width=0.12\textwidth]
  {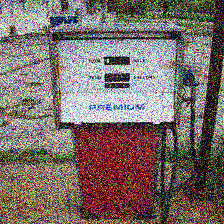} 
  \caption{Original (left) and noisy examples (middle and right) with standard deviation 0.3 (top) and 0.4 (bottom) 
  respectively. In each row, the left and right images are classified correctly while the one in the middle is classified incorrectly as church in both cases.}
  \label{fig:originals_of_fig9}
\end{figure}

\textbf{Zero-shot Classification Using ImageBind:} To verify that ImageBind can classify images accurately without any further training, we have used the publicly available ImageBind model to classify all the images in the Imagenette dataset. As described in Section \ref{sec:preli}, the classification procedure is straightforward. For each of the names of the ten classes, the model computes their representations in the shared 1,024-dimensional embedding space. Given an image, the model computes its representation using the transformer model for the image modality. The softmax is applied to the ten inner products between the image representation and the representations of the ten class names; subsequently, the image is classified into the class with the highest probability. Fig. \ref{fig:conf_matrix} shows the confusion matrices for all the images in the Imagenette dataset. Images in both the training and validation sets are included, as no further training is involved. One can see most of the classes are classified accurately with an overall accuracy of 99.38\%. Out of
13,394 images, 83 are classified incorrectly. While the results are very high, they are consistent with the pairwise cosine similarities shown in Fig. \ref{fig:similarity_dist}.

\textbf{Systematic Evaluation:} As the zero-shot classification suggests the model performs well, we explore how the model performs when it is evaluated systematically. The systematic evaluation should consider the underlying distributions in the input space. Ideally, one could generate random samples from the input space using the probability distribution defined in the input space for an application. Given the high dimensionality of the input space, it is not feasible. We have approximated it in three different ways to be effective and practical. First, we use the proposed representation matching procedure to find the ones that match the mean representations in the other nine classes for each image. While it is similar to adversarial attacks~\citep{goodfellow2015explaining,szegedy2014intriguing,chakraborty2018adversarial}, our method matches the representations, and they are agnostic to any classifier. In addition, our method finds a subspace of images whose representations are stable. Furthermore, we explore the model's behavior when adding Gaussian noise using the linear approximation discussed in Section \ref{sub:normal}.

More specifically, for each of the 13,394 images in the Imagenette dataset, we use the representation matching procedure to find one image for each of the other nine classes by matching the mean representation of the class. To ensure the mean representation is a valid one, we first compute the representations for all the images in a class, find the average of these representation vectors, and then identify the image whose representation is closest to the mean vector and use its representation as the mean representation for the class.  
We are able to match all of them with visually invisible changes.
Fig. \ref{fig:overall} and Fig. \ref{fig:overall_golf} show two typical examples.
In Fig. \ref{fig:overall}(a), a parachute image is matched to the mean representation of the tench class,
English Springer, cassette player, and chain saw. While the images are visually indistinguishable, their
representations given by the ImageBind model are very different, as seen from the low-dimensional projections of the representations. Their classification probabilities given by the {\em unmodified} ImageBind model are shown in Fig. \ref{fig:overall}(bottom); these probabilities show that these images
are classified with very high confidence to the matched class rather than the parachute class (the last column). As these images visually belong to the parachute class, the classification accuracy of the same ImageBind model is therefore 0\%.
Similarly, Fig. \ref{fig:overall_golf} shows five more examples aligned to the other five classes.
Once more, these images are visually indistinguishable, yet their representations match the other classes rather than the golf ball class; the classification accuracy of the same ImageBind model is 0\% again.
We monitor the average pixel differences when aligning a particular representation to ensure that no significant visual distortions are introduced. See Supplemental details for PSNR (peak signal-to-noise ratio) and SSIM (structural similarity index measure)  values between the original and embedding-aligned images. A typical optimization process is shown in Fig. \ref{fig:training_dynamics}, 
where the cosine similarity reaches 0.98; while the average absolute pixel difference increases, it stays small (less than 0.03) when the cosine similarity is high.
See the Supplemental materials to check the actual difference images for two typical examples.
We note that the average absolute pixel difference remains small in all the $13394 \times 9$ cases.
Based on our results, the systematic accuracy of the ImageBind model stands at 0\%, dramatically different from the very high zero-shot accuracy.

\begin{figure}[ht]
  \centering
  \includegraphics[width=0.12\textwidth]
  {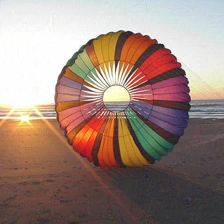}
  \includegraphics[width=0.12\textwidth]
  {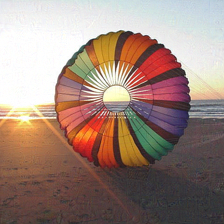}
  \includegraphics[width=0.12\textwidth]
  {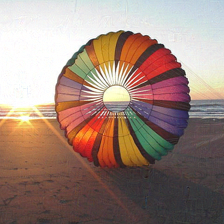}\\

  \includegraphics[width=0.12\textwidth]
  {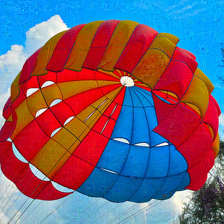}
  \includegraphics[width=0.12\textwidth]
  {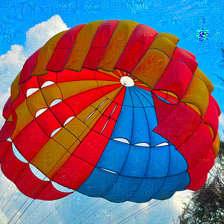}
  \includegraphics[width=0.12\textwidth]
  {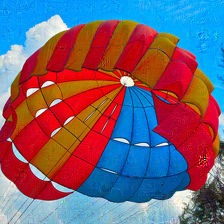}
  \caption{Different parachute images that have embedding aligned with class tench (left), English Springer (middle), and church (right); noise with a standard deviation of 0.02 is added to each of the images and they are still classified as the aligned classes respectively.}
  \label{fig:small_std}
\end{figure}


\textbf{Quantitative Evaluation:} One could argue that the images we have are essentially point-wise, meaning that we lack knowledge regarding how images in the neighborhood of a given image are classified despite our ability to match its representation. To address this issue, we have conducted two additional groups of experiments. In the first group, we take an image from the original Imagenette dataset, introduce Gaussian noise with a specified standard deviation, and then assess the classification performance provided by the ImageBind model. Note that there are infinitely many such images, and we generate 100 random images for each of the original images. Fig. \ref{fig:conf_matrix_noisy} shows the confusion matrix for 100 images, each created by adding the specified Gaussian noise to one image from each of the ten classes. We note that when the standard deviation of the noise is 0.3 or lower, all of them are classified correctly. As the standard deviation of noise increases, some of these images start to be misclassified. Several chosen examples are shown in Fig. \ref{fig:originals_of_fig9}. While these images visually belong to the same class (e.g., noisy tenches), they are classified differently. As predicted by the linear approximation model discussed in Section \ref{sub:normal}, the ImageBind is not robust to added Gaussian noise and is inconsistent when classifying such images.

Given the results we have so far, one may wonder how the ImageBind model behaves around the images that match a specified representation in the presence of Gaussian noise. To address this, we consider a set of images whose representations aligned with the given one (e.g., parachute $\rightarrow$ tench), similar to the examples shown in Fig. \ref{fig:overall} and Fig. \ref{fig:overall_golf}. Subsequently, we conduct experiments to classify these images after adding Gaussian noise with different standard deviations; 
for 0.02, all the noisy images are classified as the aligned classes (e.g., tench) with high confidence. Two examples are shown in Fig. \ref{fig:small_std}; however, for 0.03, most of them are classified as parachute. We observe similar results for other images in other classes as well. 
Note that the standard deviation is much smaller than that when we apply the same steps to the original image.
As the singular values of the Jacobians at the original and matched images differ (see Fig. \ref{fig:singulars_lc} for an example), they might play a role in the observed differences; the exact underlying mechanism for the differences is being investigated.

Fig.~\ref{fig:training_dynamics} (left) shows the evolution of loss when matching a specified target embedding. The local fluctuations are due to the inaccuracy of the gradient optimization. We use a small step size to make sure it converges.  The right panel shows that cosine similarity increases steadily. 
The algorithm is not sensitive to the learning rate and works effectively across a broad range of values, spanning from 0.001 to 0.09.  For instance, with a learning rate of 0.001, convergence is achieved in around 25,000 iterations, while 0.09 requires around 3,000 iterations. The visual differences in the resulting images are not noticeable.

\begin{figure}[ht]
  \centering
   \includegraphics[width=0.47\textwidth]{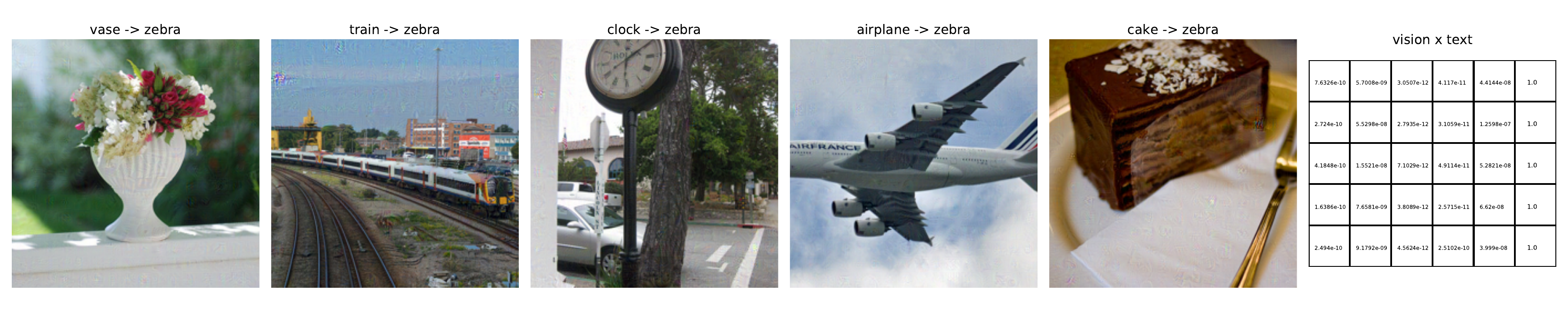}\\

   \includegraphics[width=0.47\textwidth]{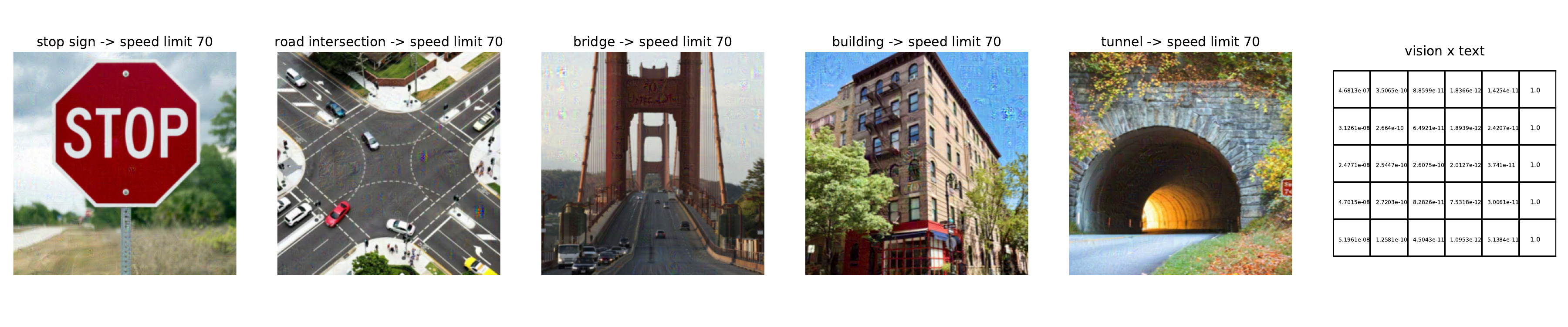}\\
  \caption{Additional examples involving MS-COCO dataset and real-world images. (top) Visually very different images from MS-COCO dataset (e.g., a vase, a train, a clock, an airplane, a cake) have very similar embeddings and are classified as zebra. The samples are randomly chosen. (bottom) Real-world scenarios where the proposed method is applied. The images are taken randomly from the web. All the matched images (for a stop sign, a road intersection, a bridge, a building, and a tunnel) are recognized as the sign ``speed limit 70". 
}
  \label{fig:overall_diff_other_datasets}
\end{figure}

\textbf{Qualitative Evaluation:} Our key result is that the semantic meanings of the embeddings given by transformer models are fundamentally limited as different inputs share similar embeddings while visually indistinguishable inputs have very different embeddings. As the techniques are model and dataset-agnostic, they should be effective on different vision-language transformer models and datasets, including ones for other modalities. To verify this, we have also conducted experiments with other vision datasets; for example, Fig.~\ref{fig:overall_diff_other_datasets} includes results on the MS-COCO dataset and real-world images. The example shows that the proposed procedure performs well in this dataset. In addition, we computed zero-shot accuracy on subsets of images from two additional datasets, alongside Imagenette and MS-COCO. Although the zero-shot accuracy aligns with the state-of-the-art for those datasets, a systematic evaluation reveals a stark contrast, resulting in 0\% accuracy. See the appendix for more detailed results with other datasets and multimodal models from HuggingFace\footnotemark\footnotetext{https://huggingface.co/docs/transformers/model\_doc/clipseg}. 

\textbf{Robust Detection of Adversarial Modifications:} As mentioned earlier, we observe that the embedding matched images exhibit much higher sensitivity to Gaussian noise compared to the corresponding original images. Leveraging this insight, we develop an algorithm to detect reliably and consistently whether an image has been modified adversarial. The process involves adding Gaussian noise of a specified standard deviation to a given image, followed by classifying both the original and the noise-added image using ImageBind. If the classification labels agree, we infer that the image remains unmodified; a discrepancy indicates modification. We apply this method to various datasets; for example, Table \ref{detection_res} summarizes the results for the Imagenette dataset. Our findings indicate that at low levels of Gaussian noise (with a small standard deviation), the detection of modified images is unreliable. However, introducing Gaussian noise with a standard deviation of 0.04 can detect them reliably. Note that a high standard deviation can misclassify some of the original images as the modified ones. Therefore, we determine that a standard deviation in the range of 0.04 to 0.3 gives accurate and reliable detection for this dataset. While the exact range of the standard deviation for added Gaussian noise may be different for a different dataset, the pattern that adversarial modifications can be detected reliably by adding Gaussian noise with a large range of standard deviation remains the same.

\begin{table}
\begin{tabularx}{0.49\textwidth} { 
  | >{\raggedright\arraybackslash}X 
  | >{\centering\arraybackslash}X
  | >{\raggedleft\arraybackslash}X
  | >{\raggedleft\arraybackslash}X
  | >{\raggedleft\arraybackslash}X
  | >{\raggedleft\arraybackslash}X
  | >{\raggedleft\arraybackslash}X
  | >{\raggedleft\arraybackslash}X | }
 \hline
 Std & 0.02 & 0.03 & 0.04 & 0.15 & 0.3 & 0.7 & 0.9\\
 \hline
 Acc  & 50\% & 64\% & 100\% & 100\% & 100\% & 81\% & 50\% \\
\hline
\end{tabularx}
\label{tab:match_success_rate}
\caption{The detection accuracy relative to the standard deviation of added Gaussian noise showcases how the precision of identifying modified images varies with the intensity of noise applied.}
\label{detection_res}
\end{table}

\section{Discussion}
By using a gradient descent procedure with mathematical analyses, we characterize the embedding space of a vision transformer locally. Note that the proposed framework can be applied to characterize any model directly, provided that the input varies continuously, allowing for accurate estimation of the Jacobian. 
One may categorize our framework as an adversarial attack technique. While our embedding matching procedure can be used to generate effective adversarial examples, it is fundamentally different. Our technique is classifier agnostic and does not exploit features specific to classifiers. Consequently, our examples with matched embeddings will appear to be the same to any classifier or downstream model that builds on embeddings. In addition, our goal is to characterize the embedding space and provide a systematic description of the space. In contrast, adversarial attack techniques find isolated examples for specific inputs. 

The paper clearly demonstrates that the currently deployed transformer-based vision-language models may have limited inherent generalization on new inputs while achieving high classification accuracy on common datasets.  
We show that images with i.i.d. Gaussian noise induce a normal distribution approximately in the representation space. The results are directly verified by experiments. For classification, a hallmark characteristic of a robust classifier is its invariance to changes within the manifold given the transformations that do not change the identities and its sensitivity to changes in the space orthogonal to the manifold. 
While much more studies need to be done to be comprehensive and systematic, our results show that transformer-based vision-language models do not exhibit the trait. Therefore, applying such models to critical applications should be evaluated extensively.

\section{Conclusion}
In this paper, we demonstrate the local structures 
of vision-language models and their importance using gradient descent and mathematical analyses. For the ImageBind model, using the entire Imagenette dataset, we show that for every image in the dataset, there is a subspace where the representations are classified with high confidence belonging to another class, even though the images are visually indistinguishable. Furthermore, we show how the samples from a normal distribution are classified using a linear approximation. Note that our method and results do not depend on the model details; therefore, we expect similar results for other transformer-based models. It is attempting to conclude that recent pretrained models can be used to build any effective application based on their zero-shot performance on benchmark datasets. While such models give impressive performance, their inherent generalization abilities might be limited by the properties of the underlying embedding spaces; therefore, such models should be evaluated systematically beyond datasets before considering their deployment in critical applications.

\begin{contributions} 
    Briefly list author contributions. 
    This is a nice way of making clear who did what and to give proper credit.
    This section is optional.

    H.~Q.~Bovik conceived the idea and wrote the paper.
    Coauthor One created the code.
    Coauthor Two created the figures.
\end{contributions}

\begin{acknowledgements} 
    Briefly acknowledge people and organizations here.

    \emph{All} acknowledgements go in this section.
\end{acknowledgements}

\bibliography{uai2024}

\begin{thebibliography}{39}
\providecommand{\natexlab}[1]{#1}
\providecommand{\url}[1]{\texttt{#1}}
\expandafter\ifx\csname urlstyle\endcsname\relax
  \providecommand{\doi}[1]{doi: #1}\else
  \providecommand{\doi}{doi: \begingroup \urlstyle{rm}\Url}\fi

\bibitem[Alayrac et~al.(2022)Alayrac, Donahue, Luc, Miech, Barr, Hasson, Lenc, Mensch, Millican, Reynolds, et~al.]{alayrac2022flamingo}
Jean-Baptiste Alayrac, Jeff Donahue, Pauline Luc, Antoine Miech, Iain Barr, Yana Hasson, Karel Lenc, Arthur Mensch, Katherine Millican, Malcolm Reynolds, et~al.
\newblock Flamingo: a visual language model for few-shot learning.
\newblock In \emph{Advances in Neural Information Processing Systems}, volume~35, pages 23716--23736, 2022.

\bibitem[Baevski et~al.(2020)Baevski, Zhou, Mohamed, and Auli]{baevski2020wav2vec}
Alexei Baevski, Yuhao Zhou, Abdelrahman Mohamed, and Michael Auli.
\newblock wav2vec 2.0: A framework for self-supervised learning of speech representations.
\newblock In \emph{Advances in Neural Information Processing Systems}, volume~33, pages 12449--12460, 2020.

\bibitem[Bhojanapalli et~al.(2021)Bhojanapalli, Chakrabarti, Glasner, Li, Unterthiner, and Veit]{bhojanapalli2021understanding}
Srinadh Bhojanapalli, Ayan Chakrabarti, Daniel Glasner, Daliang Li, Thomas Unterthiner, and Andreas Veit.
\newblock Understanding robustness of transformers for image classification.
\newblock In \emph{IEEE/CVF International Conference on Computer Vision (ICCV)}, 2021.

\bibitem[Biswas et~al.(2022)Biswas, Barao, Lazzari, McCoy, Liu, and Kostandarithes]{biswas2022geometric}
Sajib Biswas, Timothy Barao, John Lazzari, Jeret McCoy, Xiuwen Liu, and Alexander Kostandarithes.
\newblock Geometric analysis and metric learning of instruction embeddings.
\newblock In \emph{2022 International Joint Conference on Neural Networks (IJCNN)}, 2022.

\bibitem[Carlini et~al.(2023)Carlini, Others, and et~al.]{carlini2023aligned}
Nicholas Carlini, Others, and et~al.
\newblock Are aligned neural networks adversarially aligned?
\newblock \emph{arXiv preprint arXiv:2306.15447}, 2023.

\bibitem[Chakraborty et~al.(2021)Chakraborty, Alam, Dey, Chattopadhyay, and Mukhopadhyay]{chakraborty2018adversarial}
Anirban Chakraborty, Manaar Alam, Vishal Dey, Anupam Chattopadhyay, and Debdeep Mukhopadhyay.
\newblock A survey on adversarial attacks and defences.
\newblock \emph{CAAI Transactions on Intelligence Technology}, 6\penalty0 (1):\penalty0 25--45, March 2021.

\bibitem[Clark et~al.(2019)Clark, Khandelwal, Levy, and Manning]{clark-etal-2019-bert}
Kevin Clark, Urvashi Khandelwal, Omer Levy, and Christopher~D. Manning.
\newblock What does {BERT} look at? an analysis of {BERT}{'}s attention.
\newblock In \emph{Proceedings of the 2019 ACL Workshop BlackboxNLP: Analyzing and Interpreting Neural Networks for NLP}. Association for Computational Linguistics, 2019.

\bibitem[Deng et~al.(2009)Deng, Dong, Socher, Li, Li, and Fei-Fei]{imagenet2009Deng}
Jia Deng, Wei Dong, Richard Socher, Li-Jia Li, Kai Li, and Li~Fei-Fei.
\newblock Imagenet: A large-scale hierarchical image database.
\newblock In \emph{2009 IEEE Conference on Computer Vision and Pattern Recognition}, pages 248--255, 2009.

\bibitem[{Devlin} et~al.(2019){Devlin}, {Chang}, {Lee}, and {Toutanova}]{Devlin2018Bert}
Jacob {Devlin}, Ming-Wei {Chang}, Kenton {Lee}, and Kristina {Toutanova}.
\newblock {BERT: Pre-training of Deep Bidirectional Transformers for Language Understanding}.
\newblock In \emph{Proceedings of NAACL-HLT 2019}, pages 4171--4186, 2019.

\bibitem[Dosovitskiy et~al.(2021)Dosovitskiy, Beyer, Kolesnikov, Weissenborn, Zhai, Unterthiner, Dehghani, Minderer, Heigold, Gelly, Uszkoreit, and Houlsby]{dosovitskiy2021image}
Alexey Dosovitskiy, Lucas Beyer, Alexander Kolesnikov, Dirk Weissenborn, Xiaohua Zhai, Thomas Unterthiner, Mostafa Dehghani, Matthias Minderer, Georg Heigold, Sylvain Gelly, Jakob Uszkoreit, and Neil Houlsby.
\newblock An image is worth 16x16 words: Transformers for image recognition at scale.
\newblock In \emph{International Conference on Learning Representations}, 2021.

\bibitem[Emdad et~al.(2023)Emdad, Tian, Nandy, Hanna, and He]{emdad2023towards}
Forhan~Bin Emdad, Shubo Tian, Esha Nandy, Karim Hanna, and Zhe He.
\newblock Towards interpretable multimodal predictive models for early mortality prediction of hemorrhagic stroke patients.
\newblock \emph{AMIA Summits on Translational Science Proceedings}, 2023.

\bibitem[Girdhar et~al.(2023)Girdhar, El-Nouby, Liu, Singh, Alwala, Joulin, and Misra]{girdhar2023imagebind}
Rohit Girdhar, Alaaeldin El-Nouby, Zhuang Liu, Mannat Singh, Kalyan~Vasudev Alwala, Armand Joulin, and Ishan Misra.
\newblock Image{B}ind: One embedding space to bind them all.
\newblock In \emph{IEEE/CVF Conference on Computer Vision and Pattern Recognition (CVPR)}, pages 15180--15190, 2023.

\bibitem[Goodfellow et~al.(2015)Goodfellow, Shlens, and Szegedy]{goodfellow2015explaining}
Ian~J. Goodfellow, Jonathon Shlens, and Christian Szegedy.
\newblock Explaining and harnessing adversarial examples.
\newblock In \emph{International Conference on Learning Representations}, 2015.

\bibitem[Henaff(2020)]{henaff2020data}
Olivier Henaff.
\newblock Data-efficient image recognition with contrastive predictive coding.
\newblock In \emph{Proceedings of the 37th International Conference on Machine Learning}, pages 4182--4192. PMLR, 2020.

\bibitem[Herrmann et~al.(2022)Herrmann, Sargent, Jiang, Zabih, Chang, Liu, Krishnan, and Sun]{herrmann2022pyramid}
Charles Herrmann, Kyle Sargent, Lu~Jiang, Ramin Zabih, Huiwen Chang, Ce~Liu, Dilip Krishnan, and Deqing Sun.
\newblock Pyramid adversarial training improves vit performance.
\newblock \emph{arXiv preprint arXiv:2111.15121}, 2022.

\bibitem[Ilharco et~al.(2021)Ilharco, Wortsman, Wightman, Gordon, Carlini, Taori, Dave, Shankar, Namkoong, Miller, Hajishirzi, Farhadi, and Schmidt]{ilharco_gabriel_2021_5143773}
Gabriel Ilharco, Mitchell Wortsman, Ross Wightman, Cade Gordon, Nicholas Carlini, Rohan Taori, Achal Dave, Vaishaal Shankar, Hongseok Namkoong, John Miller, Hannaneh Hajishirzi, Ali Farhadi, and Ludwig Schmidt.
\newblock Openclip.
\newblock \emph{OpenCLIP}, 2021.

\bibitem[Kuznetsova et~al.(2020)Kuznetsova, Rom, Alldrin, Uijlings, Krasin, Pont-Tuset, Kamali, Popov, Malloci, Kolesnikov, Duerig, and Ferrari]{OpenImages}
Alina Kuznetsova, Hassan Rom, Neil Alldrin, Jasper Uijlings, Ivan Krasin, Jordi Pont-Tuset, Shahab Kamali, Stefan Popov, Matteo Malloci, Alexander Kolesnikov, Tom Duerig, and Vittorio Ferrari.
\newblock The open images dataset v4: Unified image classification, object detection, and visual relationship detection at scale.
\newblock \emph{IJCV}, 2020.

\bibitem[Laleh et~al.(2022)Laleh, Truhn, Veldhuizen, Han, van Treeck, Buelow, Langer, Dislich, Boor, Schulz, and Kather]{Laleh2022.03.15.484515}
Narmin~Ghaffari Laleh, Daniel Truhn, Gregory~Patrick Veldhuizen, Tianyu Han, Marko van Treeck, Roman~D. Buelow, Rupert Langer, Bastian Dislich, Peter Boor, Volkmar Schulz, and Jakob~Nikolas Kather.
\newblock Adversarial attacks and adversarial robustness in computational pathology.
\newblock \emph{Nature Communications}, 13:\penalty0 5711, 2022.

\bibitem[Li et~al.(2023)Li, Xia, Ren, Ye, Xu, and Huang]{li2023graph}
Chaoliu Li, Lianghao Xia, Xubin Ren, Yaowen Ye, Yong Xu, and Chao Huang.
\newblock Graph transformer for recommendation.
\newblock In \emph{SIGIR '23: Proceedings of the 46th Internatonal ACM SIGIR Conference on Research and Development in Information Retrieval}, pages 1680--1689, July 2023.

\bibitem[Lin et~al.(2015)Lin, Maire, Belongie, Bourdev, Girshick, Hays, Perona, Ramanan, Zitnick, and Dollár]{lin2015microsoft}
Tsung-Yi Lin, Michael Maire, Serge Belongie, Lubomir Bourdev, Ross Girshick, James Hays, Pietro Perona, Deva Ramanan, C.~Lawrence Zitnick, and Piotr Dollár.
\newblock Microsoft coco: Common objects in context.
\newblock \emph{arXiv preprint arXiv:1405.0312}, 2015.

\bibitem[Liu et~al.(2023)Liu, Li, Wu, and Lee]{llava}
Haotian Liu, Chunyuan Li, Qingyang Wu, and Yong~Jae Lee.
\newblock Visual instruction tuning.
\newblock \emph{arXiv:2304.08485}, 2023.

\bibitem[Mainuddin et~al.(2023)Mainuddin, Duan, and Dong]{mainu2023detect}
Md~Mainuddin, Zhenhai Duan, and Yingfei Dong.
\newblock Detecting compromised iot devices using autoencoders with sequential hypothesis testing.
\newblock In \emph{2023 IEEE International Conference on Big Data (BigData)}, 2023.

\bibitem[Neyshabur et~al.(2017)Neyshabur, Bhojanapalli, McAllester, and Srebro]{Neyshabur2017ExploringGI}
Behnam Neyshabur, Srinadh Bhojanapalli, David McAllester, and Nathan Srebro.
\newblock Exploring generalization in deep learning.
\newblock In \emph{NIPS}, 2017.

\bibitem[Niven and Kao(2019)]{niven-kao-2019-probing}
Timothy Niven and Hung-Yu Kao.
\newblock Probing neural network comprehension of natural language arguments.
\newblock In \emph{Proceedings of the 57th Annual Meeting of the Association for Computational Linguistics}. Association for Computational Linguistics, 2019.

\bibitem[OpenAI(2023)]{openai2023gpt4}
OpenAI.
\newblock Gpt-4 technical report.
\newblock \emph{arXiv preprint arXiv:2303.08774}, 2023.

\bibitem[Pichai(2023)]{Bard}
Sundar Pichai.
\newblock An important next step on our ai journey.
\newblock \url{https://blog.google/technology/ai/bard-google-ai-search-updates/}, 2023.

\bibitem[Qi et~al.(2023)Qi, Huang, Panda, Henderson, Wang, and Mittal]{qi2023visual}
Xiangyu Qi, Kaixuan Huang, Ashwinee Panda, Peter Henderson, Mengdi Wang, and Prateek Mittal.
\newblock Visual adversarial examples jailbreak aligned large language models.
\newblock In \emph{2nd AdvML Frontiers workshop at 40th International Conference on Machine Learning}, volume 202. PMLR, 2023.

\bibitem[Qin et~al.(2023)Qin, Zhang, Chen, Lakshminarayanan, Beutel, and Wang]{qin2023understanding}
Yao Qin, Chiyuan Zhang, Ting Chen, Balaji Lakshminarayanan, Alex Beutel, and Xuezhi Wang.
\newblock Understanding and improving robustness of vision transformers through patch-based negative augmentation.
\newblock \emph{arXiv preprint arXiv:2110.07858}, 2023.

\bibitem[Radford et~al.(2018)Radford, Narasimhan, Salimans, Sutskever, et~al.]{radford2018improving}
Alec Radford, Karthik Narasimhan, Tim Salimans, Ilya Sutskever, et~al.
\newblock Improving language understanding by generative pre-training.
\newblock \emph{OpenAI}, 2018.

\bibitem[Radford et~al.(2021)Radford, Kim, Hallacy, Ramesh, Goh, Agarwal, Sastry, Askell, Mishkin, Clark, Krueger, and Sutskever]{radford2021learning}
Alec Radford, Jong~Wook Kim, Chris Hallacy, Aditya Ramesh, Gabriel Goh, Sandhini Agarwal, Girish Sastry, Amanda Askell, Pamela Mishkin, Jack Clark, Gretchen Krueger, and Ilya Sutskever.
\newblock Learning transferable visual models from natural language supervision.
\newblock In \emph{International Conference on Machine Learning}, volume 139. PMLR, 2021.

\bibitem[Raffel et~al.(2020)Raffel, Shazeer, Roberts, Lee, Narang, Matena, Zhou, Li, and Liu]{colin2020t5}
Colin Raffel, Noam Shazeer, Adam Roberts, Katherine Lee, Sharan Narang, Michael Matena, Yanqi Zhou, Wei Li, and Peter~J. Liu.
\newblock Exploring the limits of transfer learning with a unified text-to-text transformer.
\newblock \emph{Journal of Machine Learning Research}, 21\penalty0 (140):\penalty0 1--67, 2020.

\bibitem[Salman et~al.(2024)Salman, Shams, and Liu]{salman2024intriguing}
Shaeke Salman, Md~Montasir~Bin Shams, and Xiuwen Liu.
\newblock Intriguing equivalence structures of the embedding space of vision transformers.
\newblock \emph{arXiv preprint arXiv:2401.15568}, 2024.

\bibitem[Szegedy et~al.(2014)Szegedy, Zaremba, Sutskever, Bruna, Erhan, Goodfellow, and Fergus]{szegedy2014intriguing}
Christian Szegedy, Wojciech Zaremba, Ilya Sutskever, Joan Bruna, Dumitru Erhan, Ian Goodfellow, and Rob Fergus.
\newblock Intriguing properties of neural networks.
\newblock \emph{arXiv preprint arXiv:1312.6199}, 2014.

\bibitem[Vaswani et~al.(2017)Vaswani, Shazeer, Parmar, Uszkoreit, Jones, Gomez, Kaiser, and Polosukhin]{vaswani2023attention}
Ashish Vaswani, Noam Shazeer, Niki Parmar, Jakob Uszkoreit, Llion Jones, Aidan~N. Gomez, Lukasz Kaiser, and Illia Polosukhin.
\newblock Attention is all you need.
\newblock In \emph{Advances in Neural Information Processing Systems}, 2017.

\bibitem[Xu et~al.(2023)Xu, Zhu, and Clifton]{xu2023multimodal}
Peng Xu, Xiatian Zhu, and David~A Clifton.
\newblock Multimodal learning with transformers: A survey.
\newblock \emph{IEEE Transactions on Pattern Analysis and Machine Intelligence}, 2023.

\bibitem[Zhang et~al.(2017{\natexlab{a}})Zhang, Bengio, Hardt, Recht, and Vinyals]{Zhang2016UnderstandingDL}
Chiyuan Zhang, Samy Bengio, Moritz Hardt, Benjamin Recht, and Oriol Vinyals.
\newblock Understanding deep learning requires rethinking generalization.
\newblock In \emph{International Conference on Learning Representations}, 2017{\natexlab{a}}.

\bibitem[Zhang et~al.(2017{\natexlab{b}})Zhang, Bengio, Hardt, Recht, and Vinyals]{zhang2017understanding}
Chiyuan Zhang, Samy Bengio, Moritz Hardt, Benjamin Recht, and Oriol Vinyals.
\newblock Understanding deep learning requires rethinking generalization.
\newblock \emph{arXiv preprint arXiv:1611.03530}, 2017{\natexlab{b}}.

\bibitem[Zhu et~al.(2023)Zhu, Chen, Shen, Li, and Elhoseiny]{zhu2023minigpt}
Deyao Zhu, Jun Chen, Xiaoqian Shen, Xiang Li, and Mohamed Elhoseiny.
\newblock Mini{GPT}-4: Enhancing vision-language understanding with advanced large language models.
\newblock \emph{arXiv preprint arXiv:2304.10592}, 2023.

\bibitem[Zou et~al.(2023)Zou, Wang, Kolter, and Fredrikson]{zou2023universal}
Andy Zou, Zifan Wang, J~Zico Kolter, and Matt Fredrikson.
\newblock Universal and transferable adversarial attacks on aligned language models.
\newblock \emph{arXiv preprint arXiv:2307.15043}, 2023.

\end{thebibliography}

\newpage

\onecolumn



\appendix

\section{Appendix}

\subsection{More on Vision Transformers}

Very recently, several multi-modal models have been introduced [Xu et al., 2023; Zhu et al., 2023; OpenAI, 2023; Girdhar et al., 2023].

\begin{figure}[H]
  \centering
    \includegraphics[width=0.40\columnwidth]{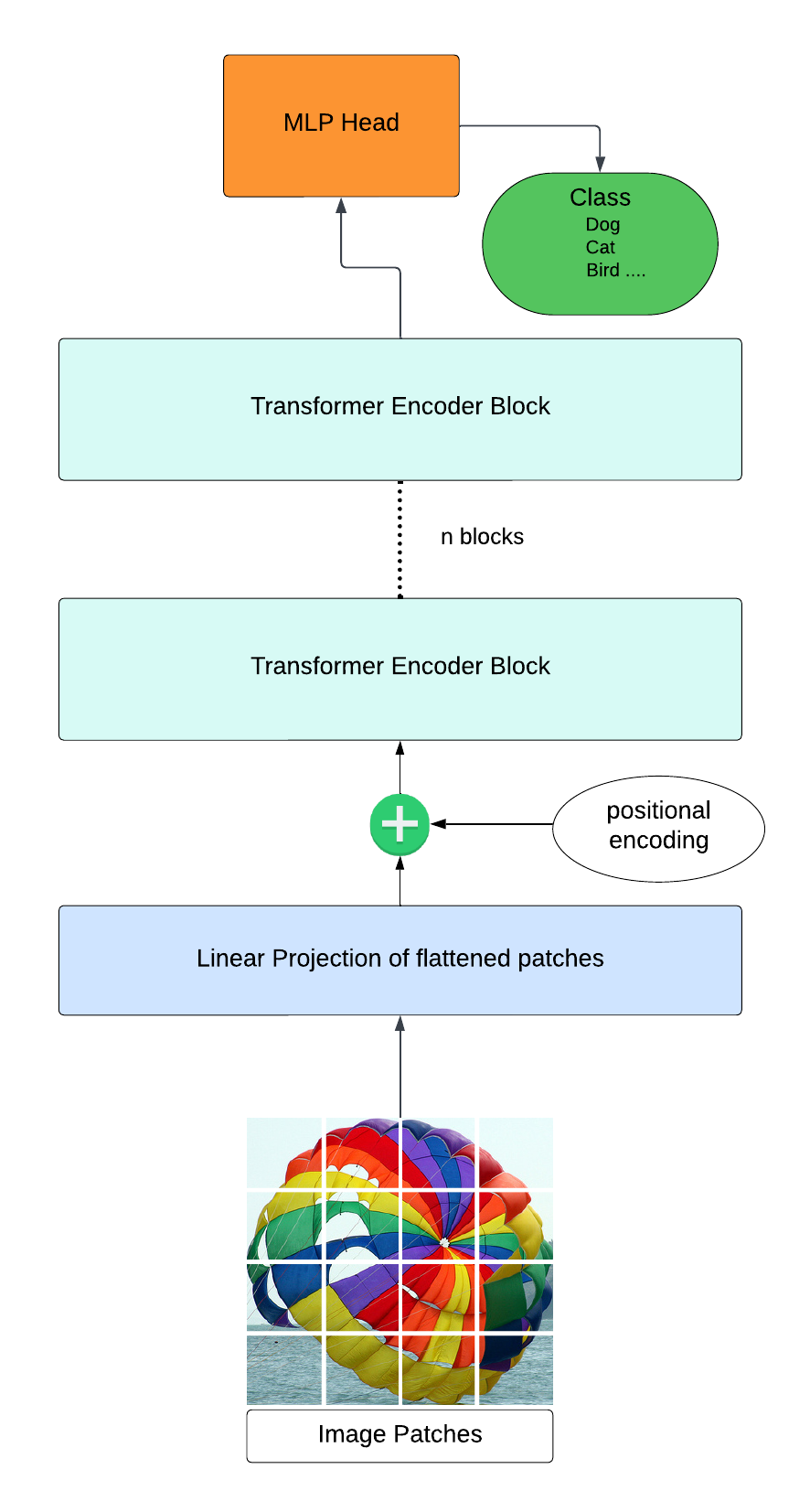}
  \caption{Vision Transformer (ViT) architecture.}
  \label{fig:vt_arch}
\end{figure}

By using a shared embedding space among different modalities, such joint models have shown to have advantages. Vision transformers have been successful in various vision tasks due to their ability to treat an image as a sequence of patches and utilize self-attention mechanisms.

A collection of transformer blocks makes up the Vision Transformer Architecture. Each transformer block comprises two sub-layers: a multi-headed self-attention layer and a feed-forward layer. The self-attention layer computes attention weights for each pixel in the image based on its relationship with all other pixels, while the feed-forward layer applies a non-linear transformation to the self-attention layer's output. The patch embedding layer separates the image into fixed-size patches before mapping each patch to a high-dimensional vector representation. These patch embeddings are then supplied into the transformer blocks to be processed further [Dosovitskiy et al., 2021].

\subsection{Additional Results}

\begin{figure}[ht]
    \centering
    \vspace{-0.10in}
\begin{tabular}{>{\centering\arraybackslash}m{.08\textwidth}m{.35in}>{\centering\arraybackslash}m{.09\textwidth}m{.05in}>{\centering\arraybackslash}m{.1\textwidth}}
    \centering\arraybackslash
    \includegraphics[width=.10\textwidth]{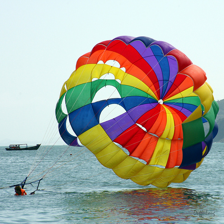} &%
    \centering\arraybackslash%
$\ +\ .01\ \times$ &%
    \includegraphics[width=.10\textwidth]{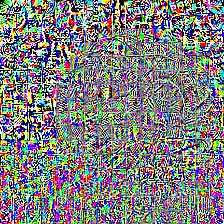} &%
    $=$ & %
    \includegraphics[width=.10\textwidth]{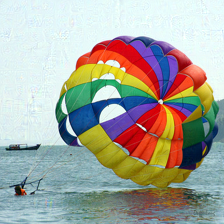} \\

    \centering\arraybackslash
    \includegraphics[width=.10\textwidth]{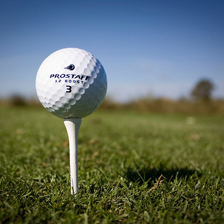} &%
    \centering\arraybackslash%
$\ +\ .01\ \times$ &%
    \includegraphics[width=.10\textwidth]{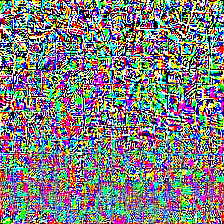} &%
    $=$ & %
    \includegraphics[width=.10\textwidth]{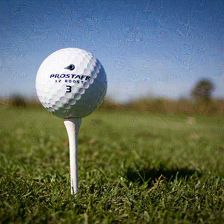} \\

    \centering\arraybackslash
    \includegraphics[width=.10\textwidth]{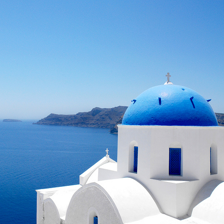} &%
    \centering\arraybackslash%
$\ +\ .01\ \times$ &%
    \includegraphics[width=.10\textwidth]{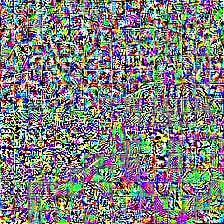} &%
    $=$ & %
   \includegraphics[width=.10\textwidth]{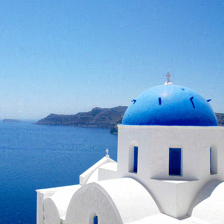} \\
\end{tabular}
\vspace{-0.10in}
    \caption{
  Pixel differences between the two images in each of the pairs; the difference images are multiplied by 100 with zero mapped to the middle value for visualization; for each pair, the left image is embedding aligned
  to another class (tench, English Springer, and tench 
  respectively from top to bottom).
    }
\label{fig:diffnoise}
\vspace{-0.20in}
\end{figure}
Here we provide more details and additional information about the results we have included in the main text.

\begin{figure*}[ht]
  \centering
  \includegraphics[width=1.0\textwidth]{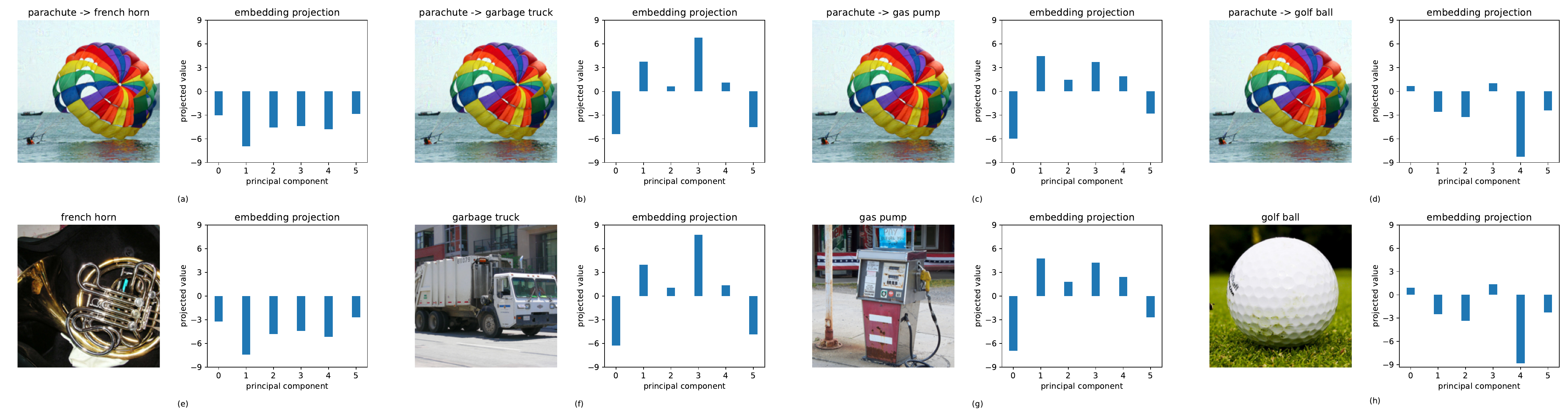}\label{fig:for_1}\\

  \caption{Same as Fig. 1, but shown while embedding-aligned to each of the last four classes instead of the first four as shown in Fig. 1.
}
  \label{fig:for_fig_1}
\end{figure*}

\begin{figure*}[ht]
  \centering

  \includegraphics[width=1.0\textwidth]{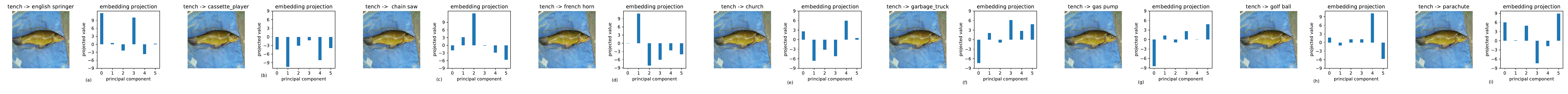}\label{fig:for_more0} \\

 \includegraphics[width=1.0\textwidth]{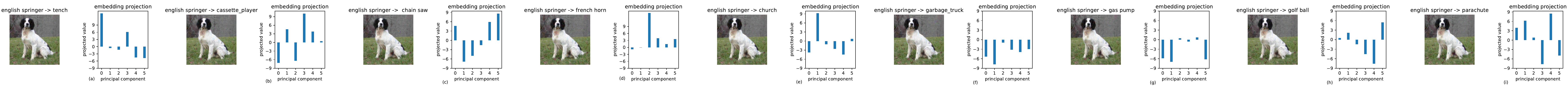}\label{fig:for_more1} \\

  \includegraphics[width=1.0\textwidth]{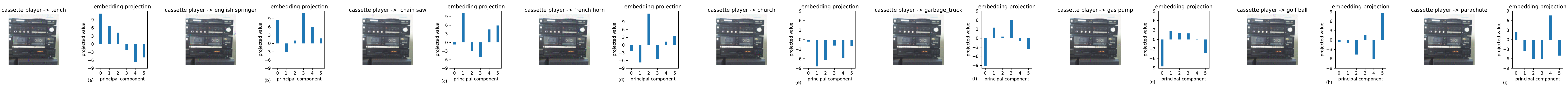}\label{fig:for_more2} \\

 \includegraphics[width=1.0\textwidth]{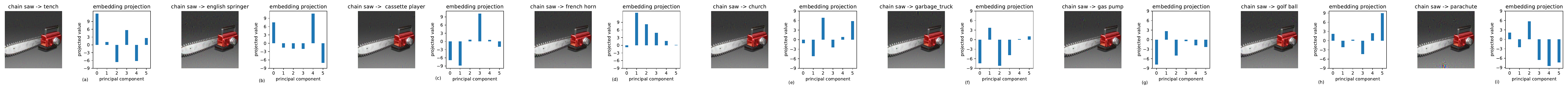}\label{fig:for_more3} \\

  \includegraphics[width=1.0\textwidth]{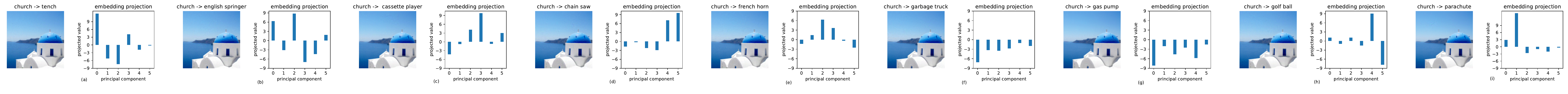}\label{fig:for_more4} \\

  \includegraphics[width=1.0\textwidth]{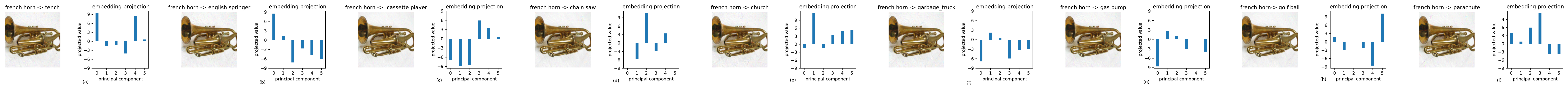}\label{fig:for_more5} \\

  \includegraphics[width=1.0\textwidth]{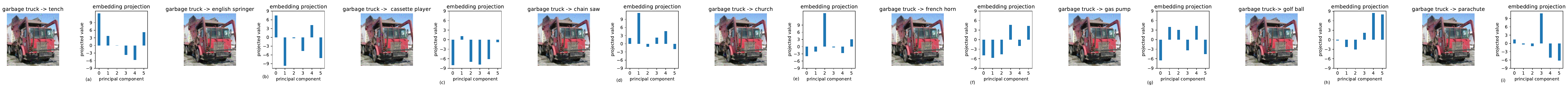}\label{fig:for_more6} \\

  \includegraphics[width=1.0\textwidth]{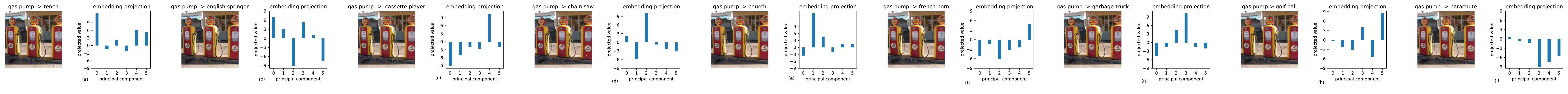}\label{fig:for_more7} \\

  \includegraphics[width=1.0\textwidth]{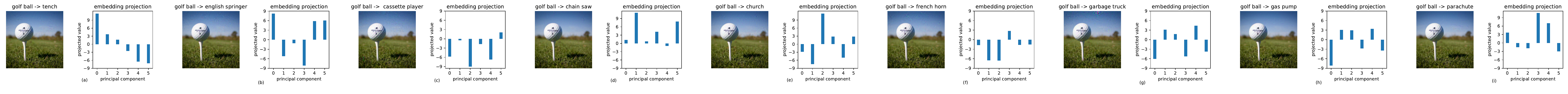}\label{fig:for_more8} \\

  \caption{All nine images and projections for all the other nine classes, from class 0 to class 8 respectively (from top to bottom). Please note that here we haven't included the embedding projections for aligned classes as in Fig. 1. 
}
  \label{fig:for_fig_1_more}
\end{figure*}

\begin{figure*}[ht]
  \centering

  \includegraphics[width=0.99\textwidth]{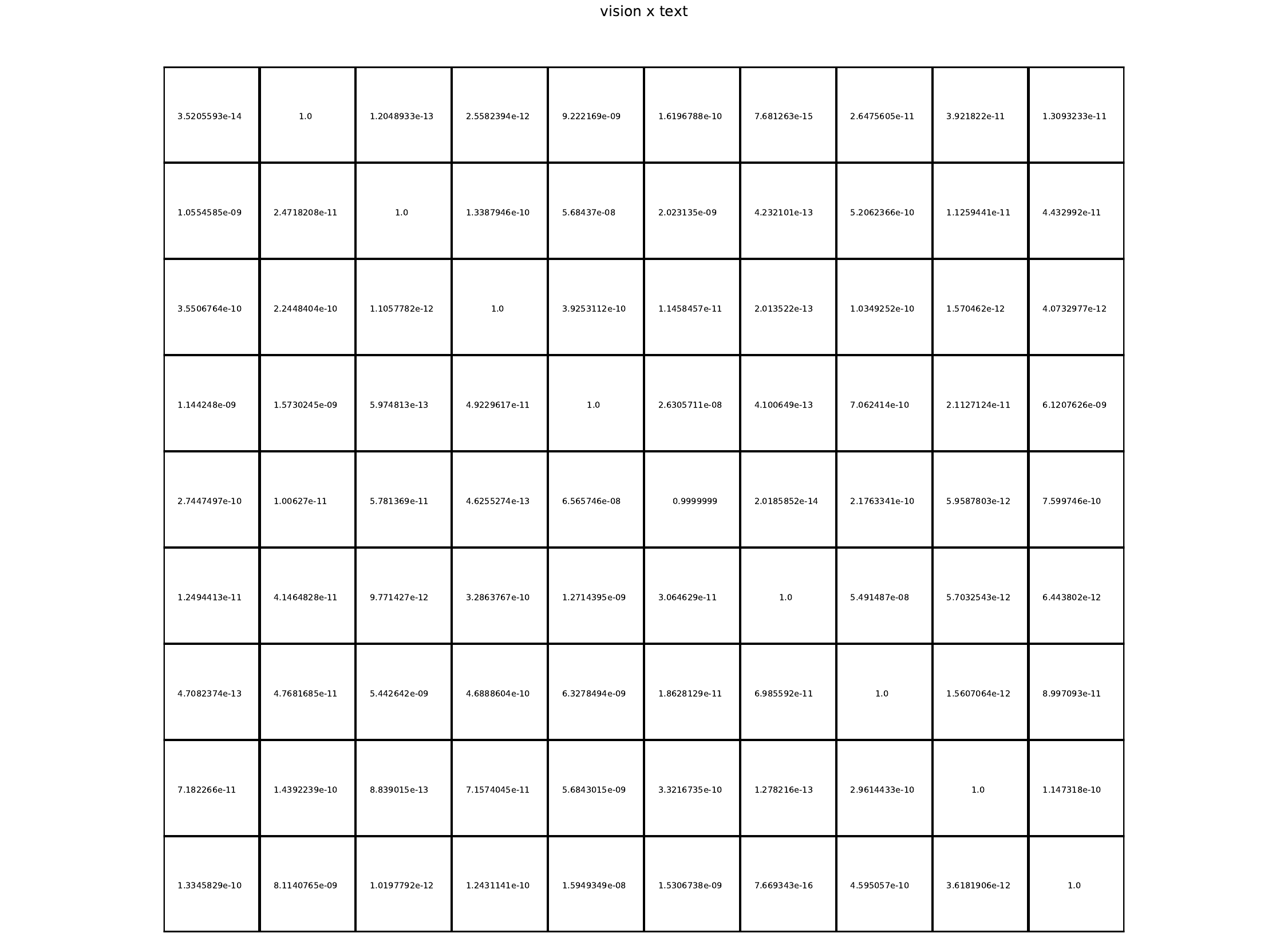}
  \caption{The full $vision \times text$ matrix with unclipped outputs for class 0 (tench), obtained using multimodal ImageBind pretrained model.}
  
  \label{fig:unclipped_0}
\end{figure*}

\begin{figure*}[ht]
  \centering

  \includegraphics[width=0.99\textwidth]{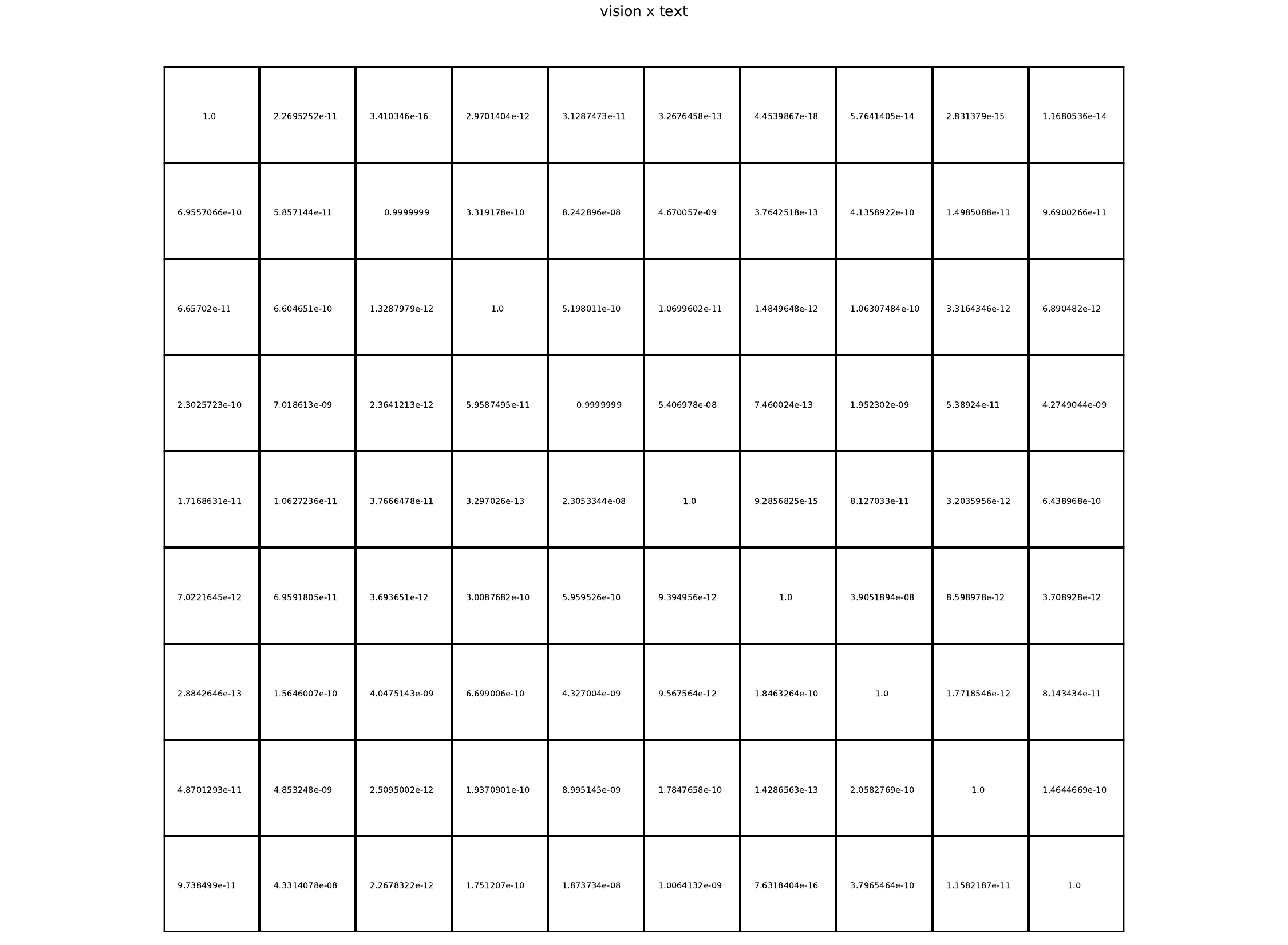}
  \caption{The full $vision \times text$ matrix with unclipped outputs for class 1 (english springer), obtained using multimodal pretrained ImageBind model.}
  
  \label{fig:unclipped_1}
\end{figure*}

\begin{figure*}[ht]
  \centering

  \includegraphics[width=0.99\textwidth]{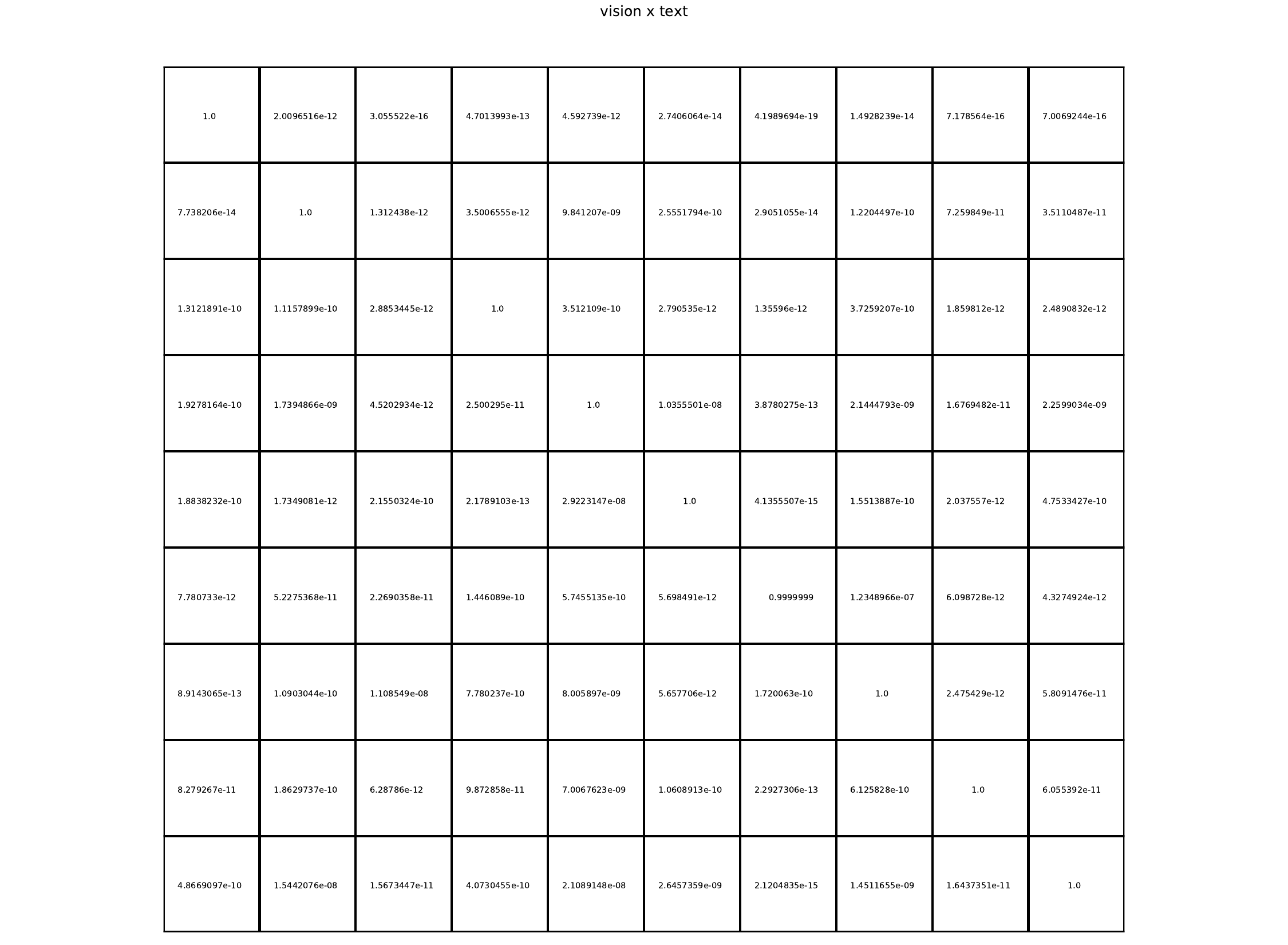}
  \caption{The full $vision \times text$ matrix with unclipped outputs for class 2 (cassette player), obtained using multimodal ImageBind pretrained model.}
  
  \label{fig:unclipped_2}
\end{figure*}

\begin{figure*}[ht]
  \centering

  \includegraphics[width=0.99\textwidth]{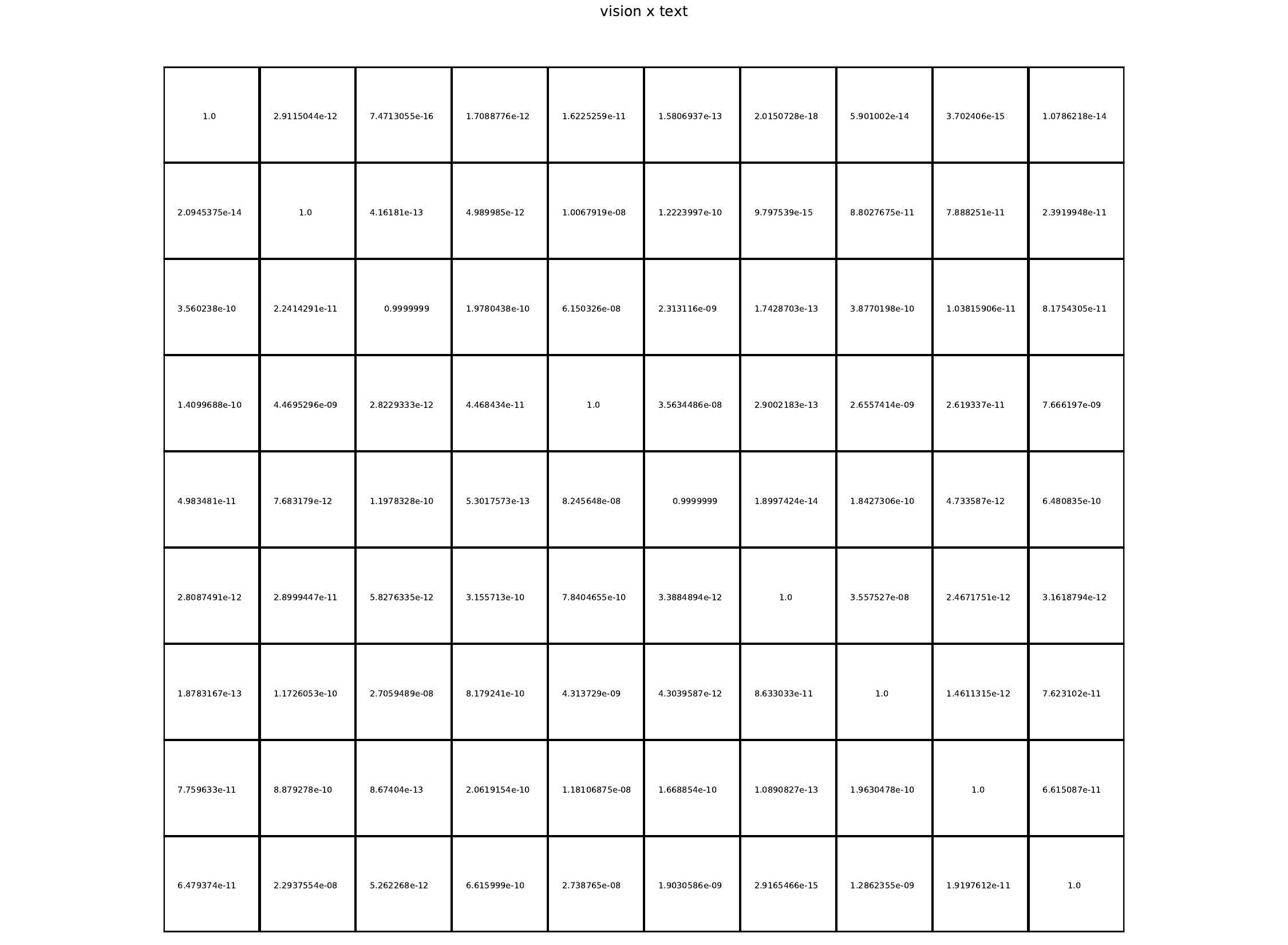}
  \caption{The full $vision \times text$ matrix with unclipped outputs for class 3 (chain saw), obtained using multimodal ImageBind pretrained model.}
  
  \label{fig:unclipped_3}
\end{figure*}

\begin{figure*}[ht]
  \centering

  \includegraphics[width=0.99\textwidth]{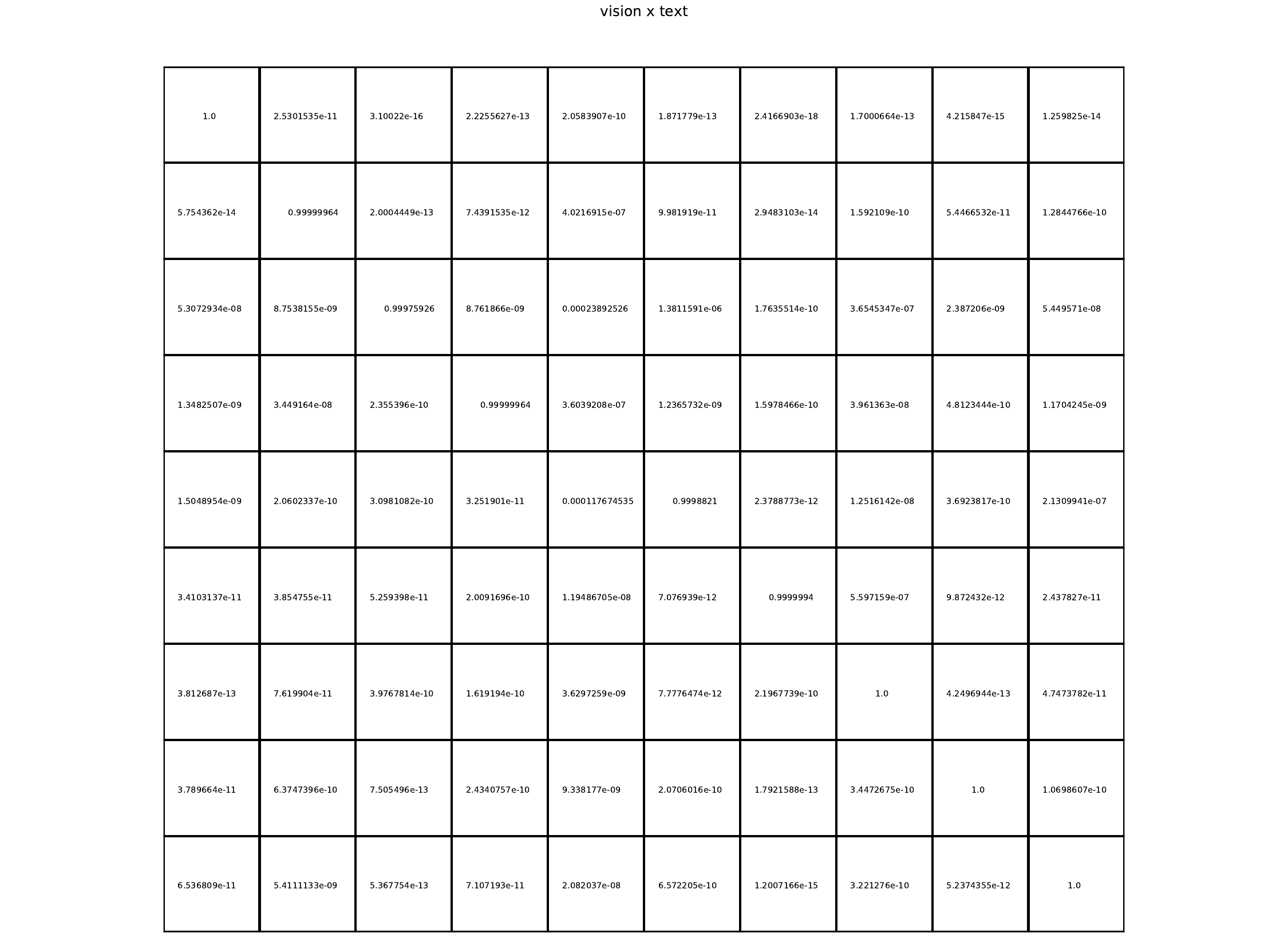}
  \caption{The full $vision \times text$ matrix with unclipped outputs for class 4 (church), obtained using multimodal pretrained ImageBind model.}
  
  \label{fig:unclipped_4}
\end{figure*}

\begin{figure*}[ht]
  \centering

  \includegraphics[width=0.99\textwidth]{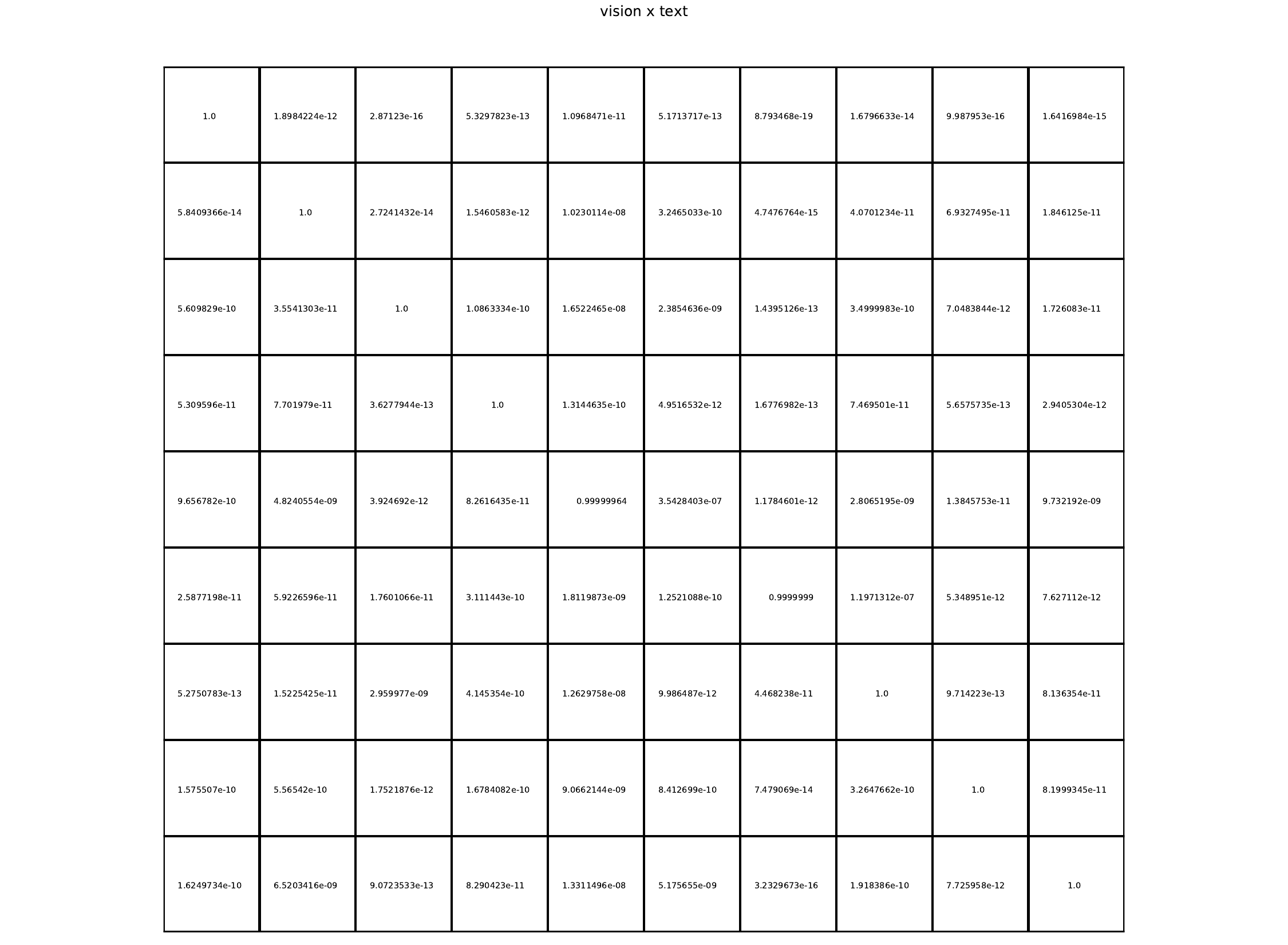}
  \caption{The full $vision \times text$ matrix with unclipped outputs for class 5 (french horn), obtained using multimodal ImageBind pretrained model.}
  
  \label{fig:unclipped_5}
\end{figure*}

\begin{figure*}[ht]
  \centering

  \includegraphics[width=0.99\textwidth]{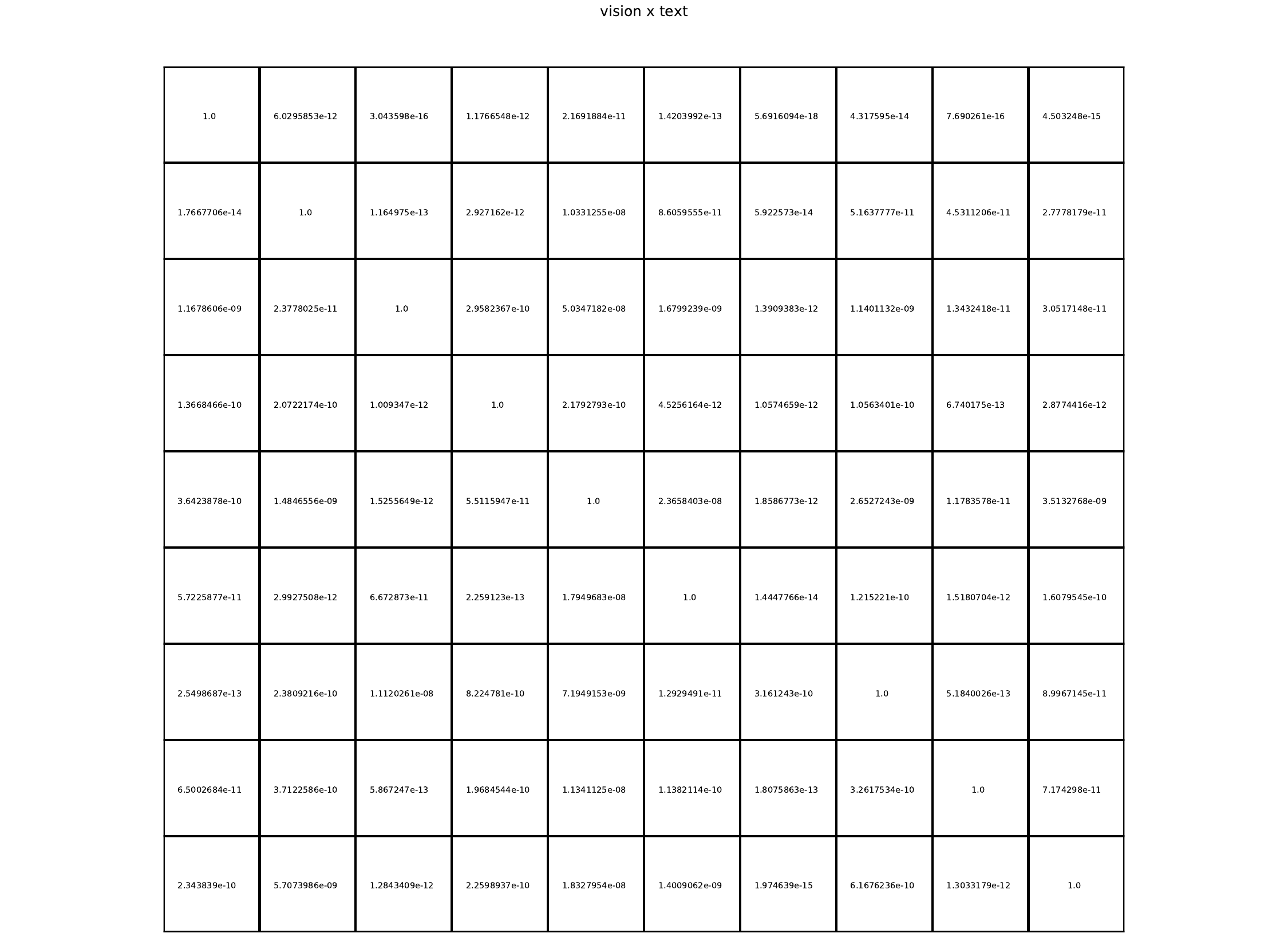}
  \caption{The full $vision \times text$ matrix with unclipped outputs for class 6 (garbage truck), obtained using multimodal ImageBind pretrained model.}
  
  \label{fig:unclipped_6}
\end{figure*}

\begin{figure*}[ht]
  \centering

  \includegraphics[width=0.99\textwidth]{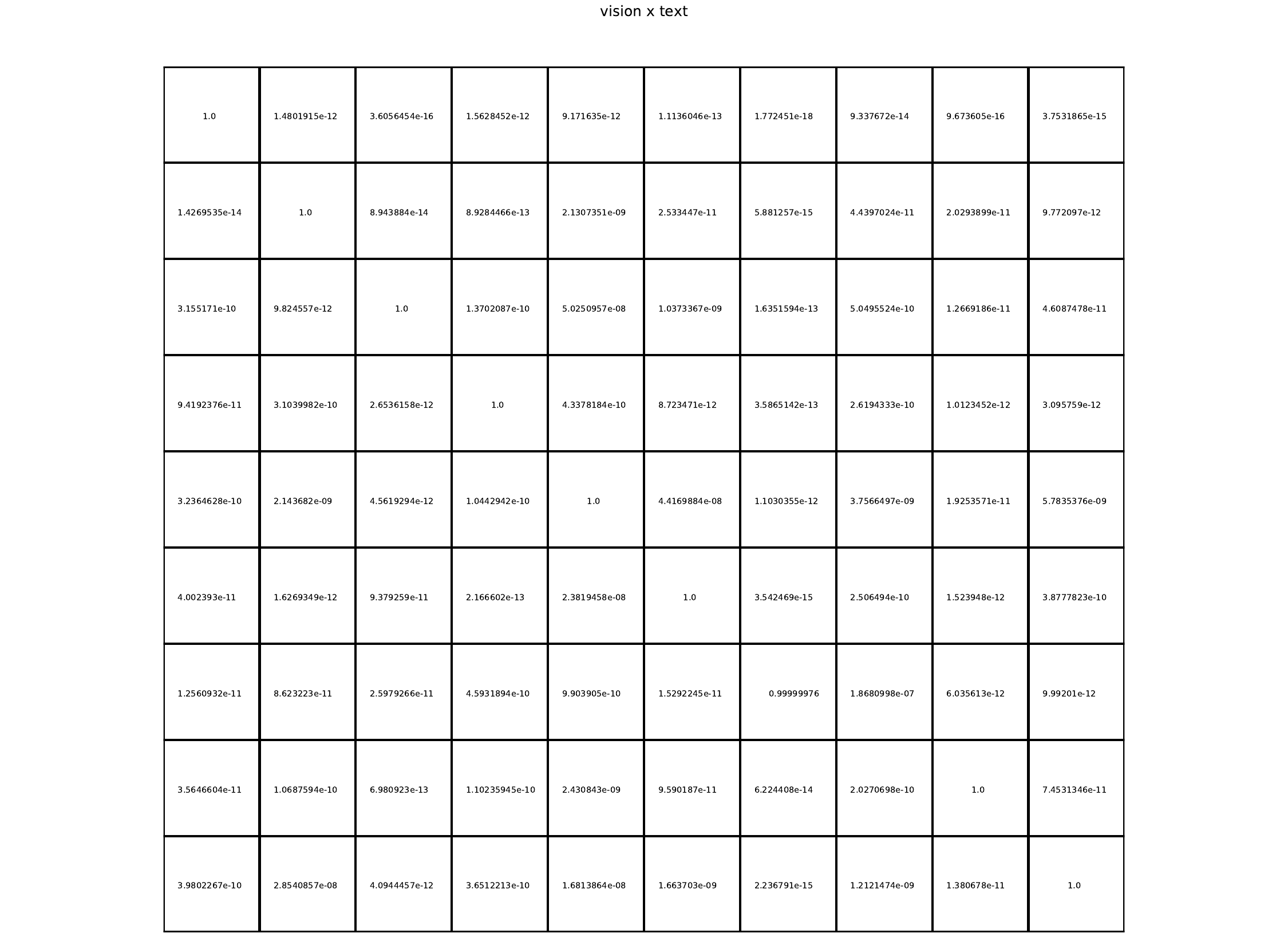}
  \caption{The full $vision \times text$ matrix with unclipped outputs for class 7 (gas pump), obtained using multimodal ImageBind pretrained model.}
  
  \label{fig:unclipped_7}
\end{figure*}

\begin{figure*}[ht]
  \centering
  \includegraphics[width=0.99\textwidth]{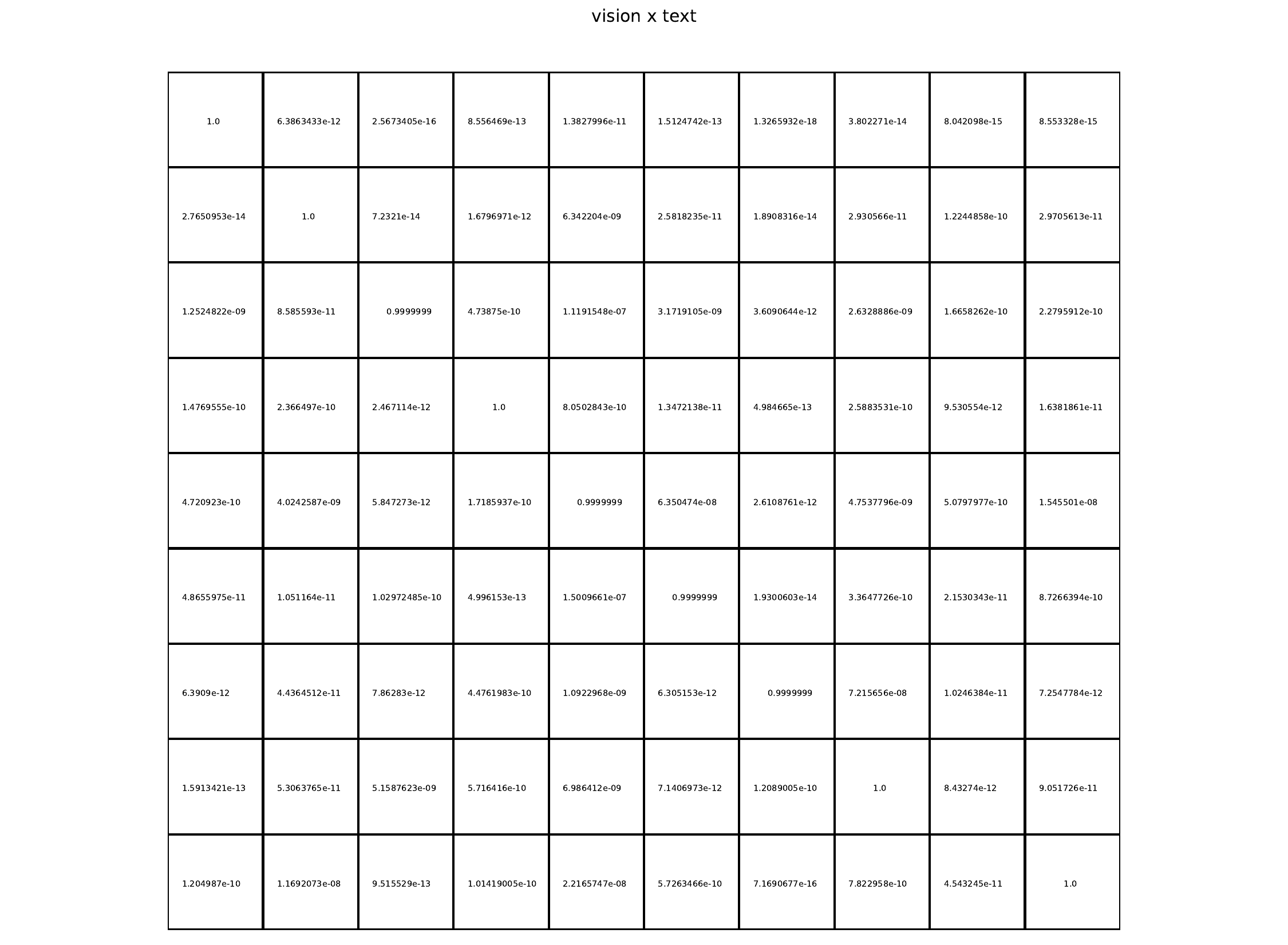}
  \caption{The full $vision \times text$ matrix with unclipped outputs for class 8 (golf ball), obtained using multimodal ImageBind pretrained model.}
  
  \label{fig:unclipped_8}
\end{figure*}

\begin{figure*}[ht]
  \centering
  \includegraphics[width=0.99\textwidth]{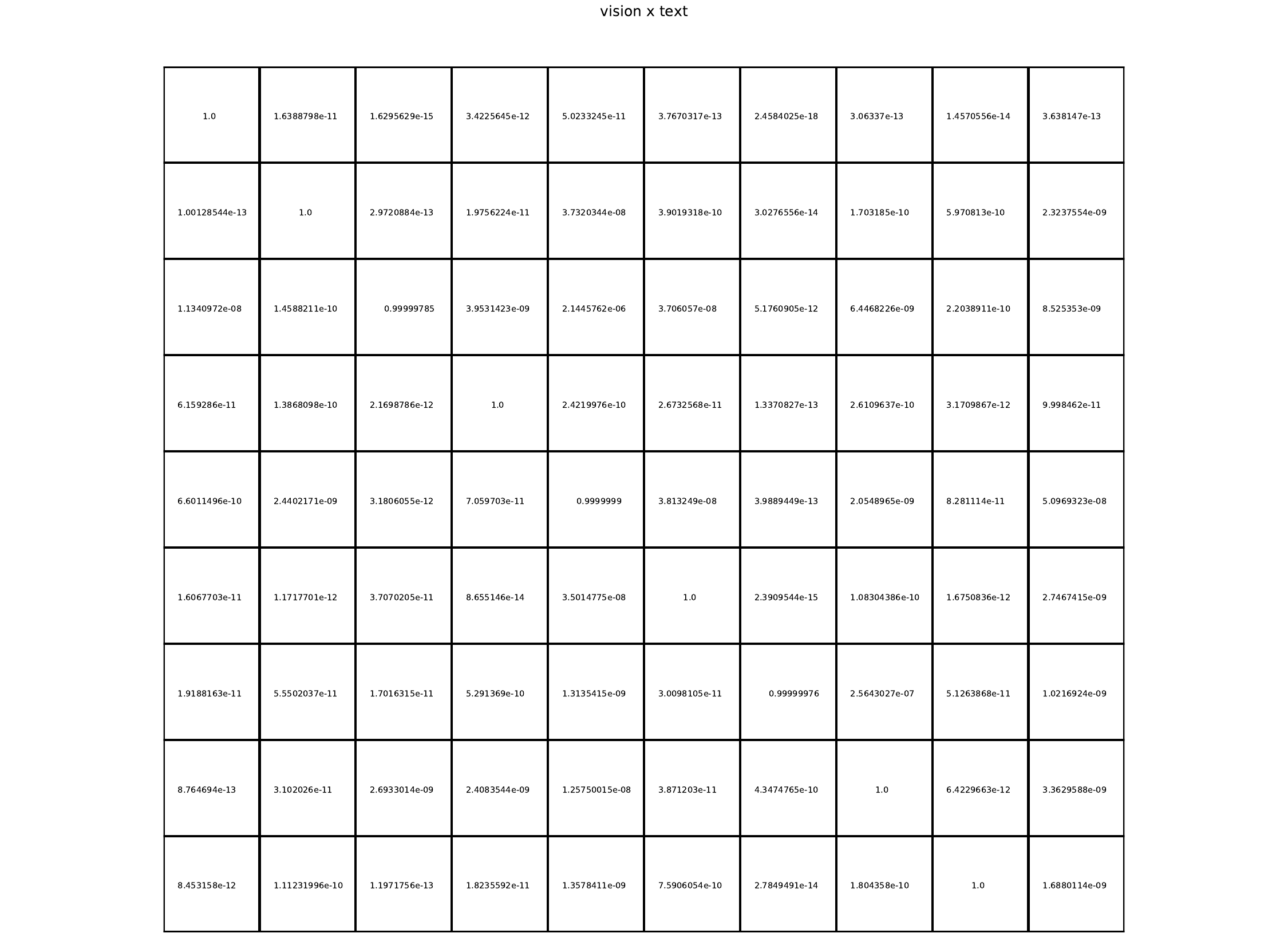}
  \caption{The full $vision \times text$ matrix with unclipped outputs for class 9 (parachute), obtained using multimodal ImageBind pretrained model.}
  
  \label{fig:unclipped_9}
\end{figure*}

\textbf{Additional Results with Imagenette Dataset:} Fig. \ref{fig:for_fig_1} shows four visually indistinguishable parachute images having very different representations as presented by their low-dimensional projections; same as Fig. 1 in the main paper, but shown while embedding-aligned to each of the last four classes of Imagenette dataset instead of the first four as shown in Fig. 1. In addition, Fig. \ref{fig:for_fig_1_more} gives a comprehensive view, consisting of all the nine embedding-aligned images and projections for all the other nine classes of Imagenette. The generation of indistinguishable images for a particular class involves aligning the representation of an image originally belonging to that specific class with representations from other classes. Fig.~\ref{fig:unclipped_0} to Fig.~\ref{fig:unclipped_9} show the classification outcomes for all these images (from class 0 to class 9 respectively), obtained from the multimodal ImageBind pretrained model used directly with no modifications. It is clear that values are either very close to 1 or very close to 0, demonstrating that the classification results are stable. Note that for all these matrices, the rows (from top to bottom) show the classification of the images from left to right respectively; therefore, each row corresponds to one image (from left to right). The columns from left to right are tench, English springer, cassette player, chain saw, church, French horn, garbage truck, gas pump, golf ball, and parachute respectively as the matrices shown in the main paper. 

\textbf{Additional Results with Other Models and Datasets:} We show that our framework exhibits versatility, being agnostic to both the model architecture and dataset characteristics. The efficacy of the procedure is model-agnostic. To substantiate and confirm this, Fig.  \ref{fig:overall_golf_clipseg} shows an example of a different multimodal model using CLIPSeg, showcasing the consistent application and effectiveness of the approach irrespective of the specific model employed. 

\begin{figure*}[ht]
  \centering

  \includegraphics[width=0.85\textwidth]{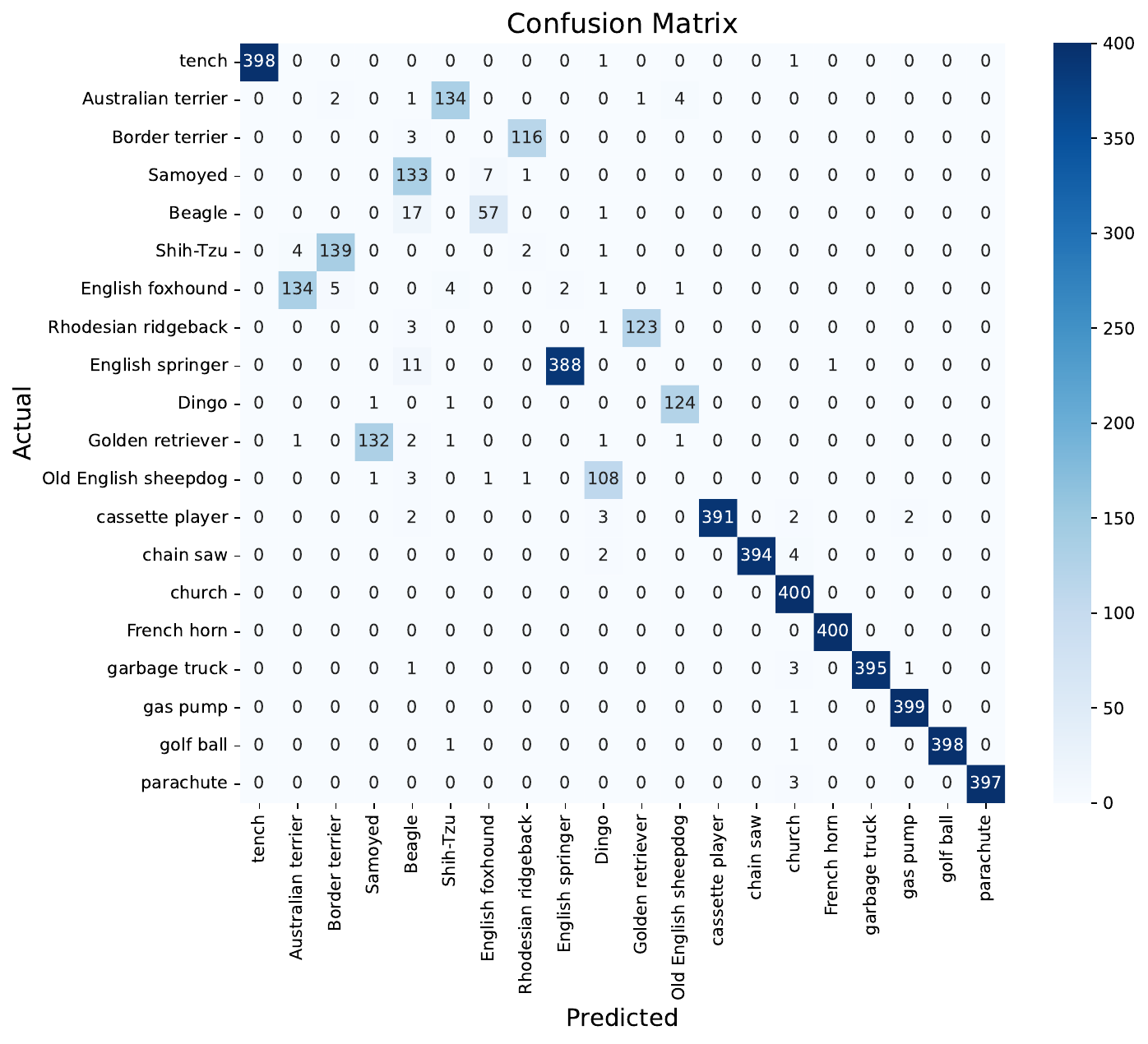}

  \caption{Confusion matrix of zero-shot classification performance for all the classes of Imagewang dataset. The overall accuracy is $75.37\%$.
  }
  \label{fig:conf_matrix_image_wang}
\end{figure*}

To maintain experimental simplicity, we focus on a subset of images extracted from both the MS-COCO dataset and the Google Open Images dataset. FiftyOne Dataset Zoo\footnotemark\footnotetext{https://docs.voxel51.com/user\_guide/dataset\_zoo/index.html} has been used to download a subset of images from both of the datasets.
\begin{figure*}[ht]
  \centering

  \includegraphics[width=0.85\textwidth]{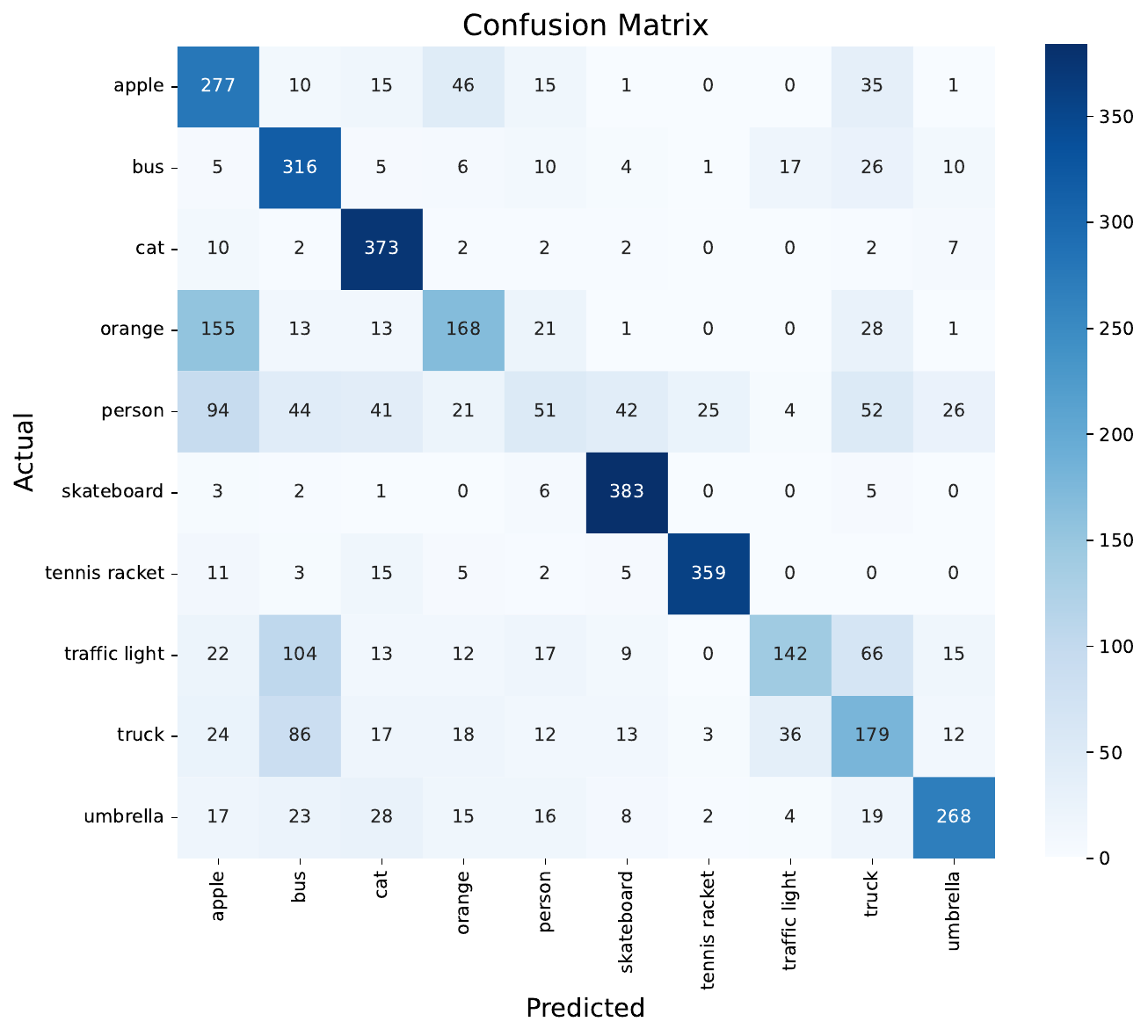}

  \caption{Confusion matrix of zero-shot classification performance for 10 classes of MS-COCO  dataset. The overall accuracy is $62.9\%$.
  }
  \label{fig:conf_matrix_MS_coco}
\end{figure*}

\begin{figure*}[ht]
  \centering

  \includegraphics[width=0.85\textwidth]{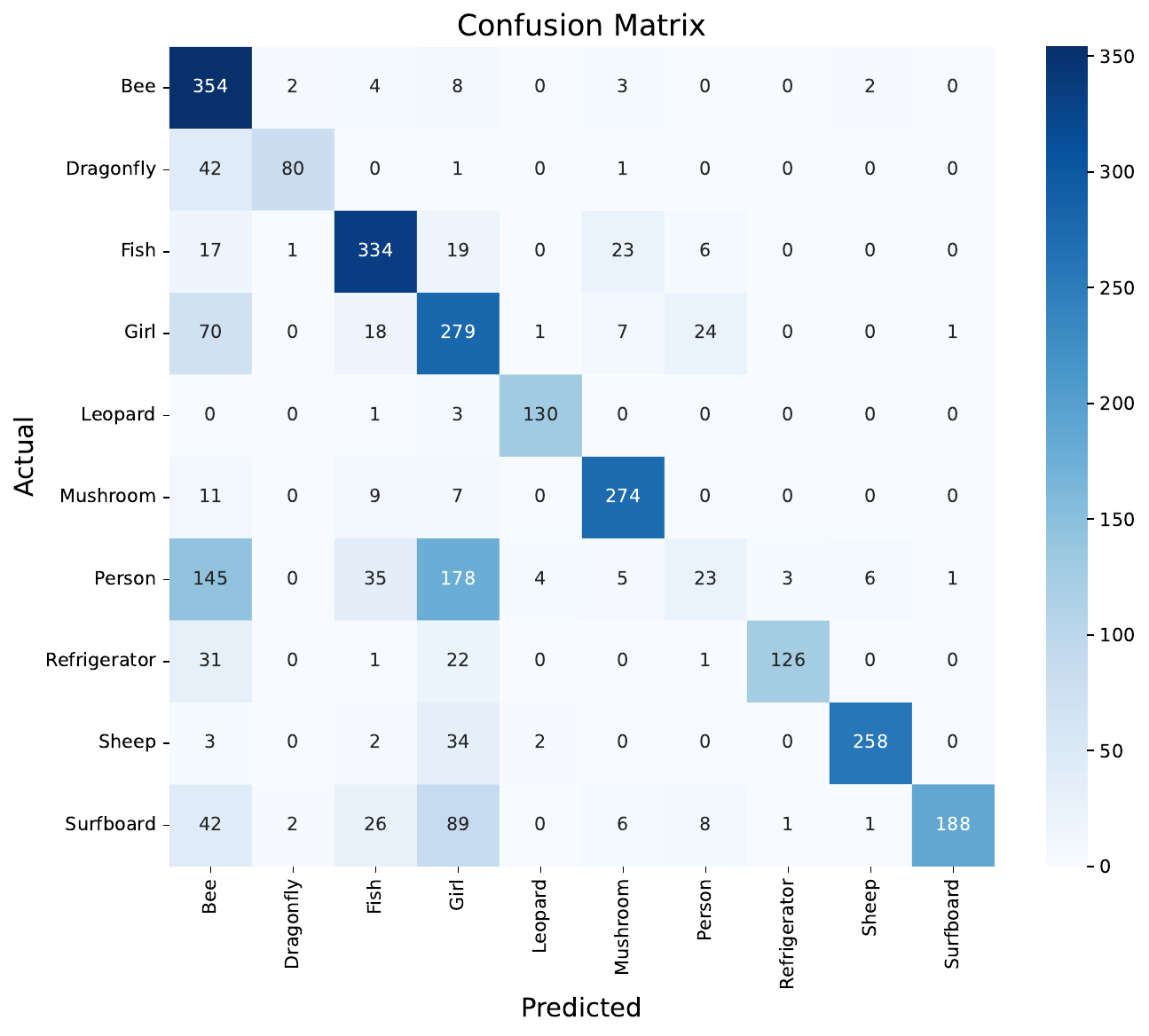}

  \caption{Confusion matrix of zero-shot classification performance for 10 classes of Google Open Images dataset. The overall accuracy is $68.88\%$.
  }
  \label{fig:conf_matrix_open-images}
\end{figure*}

\begin{table}
  \centering
    \small
  \begin{tabular}{lll}
    \toprule
    Image & Mean PSNR & Mean SSIM\\
    \midrule
    ImageNet & 47 dB & 0.982  \\
    MS-COCO & 46 dB & 0.980 \\
    Open Images & 49 dB & 0.985 \\
    
    \bottomrule
  \end{tabular}
  \label{tab:psnr_ssim}
  \caption{The average PSNR value and SSIM index between the original and embedding-aligned images to all the target images embeddings in each of the datasets; the average is computed based on 1000 examples for each dataset and the images are strictly randomly chosen from the datasets with no postselection.
  }
  \label{psnr_ssim_res}
\end{table}

\begin{figure*}[ht]
  \centering
  \includegraphics[width=1.0\textwidth]{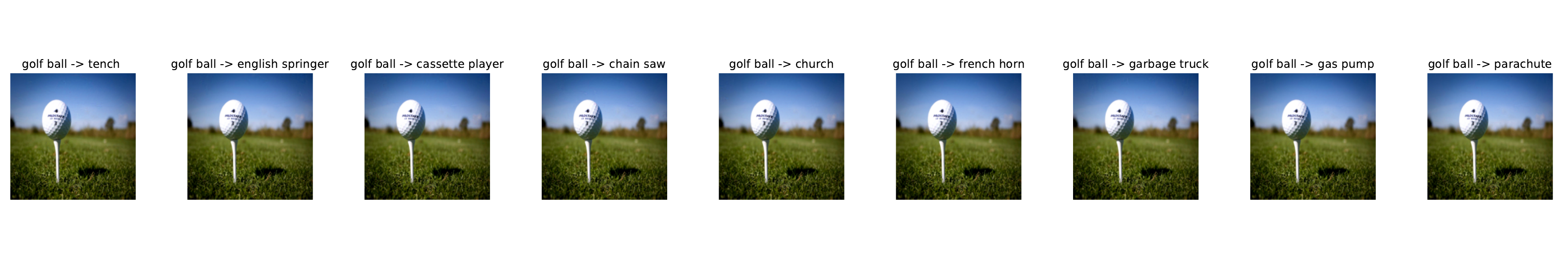}\label{fig:more_matched_clipseg}
  \includegraphics[width=1.0\textwidth]{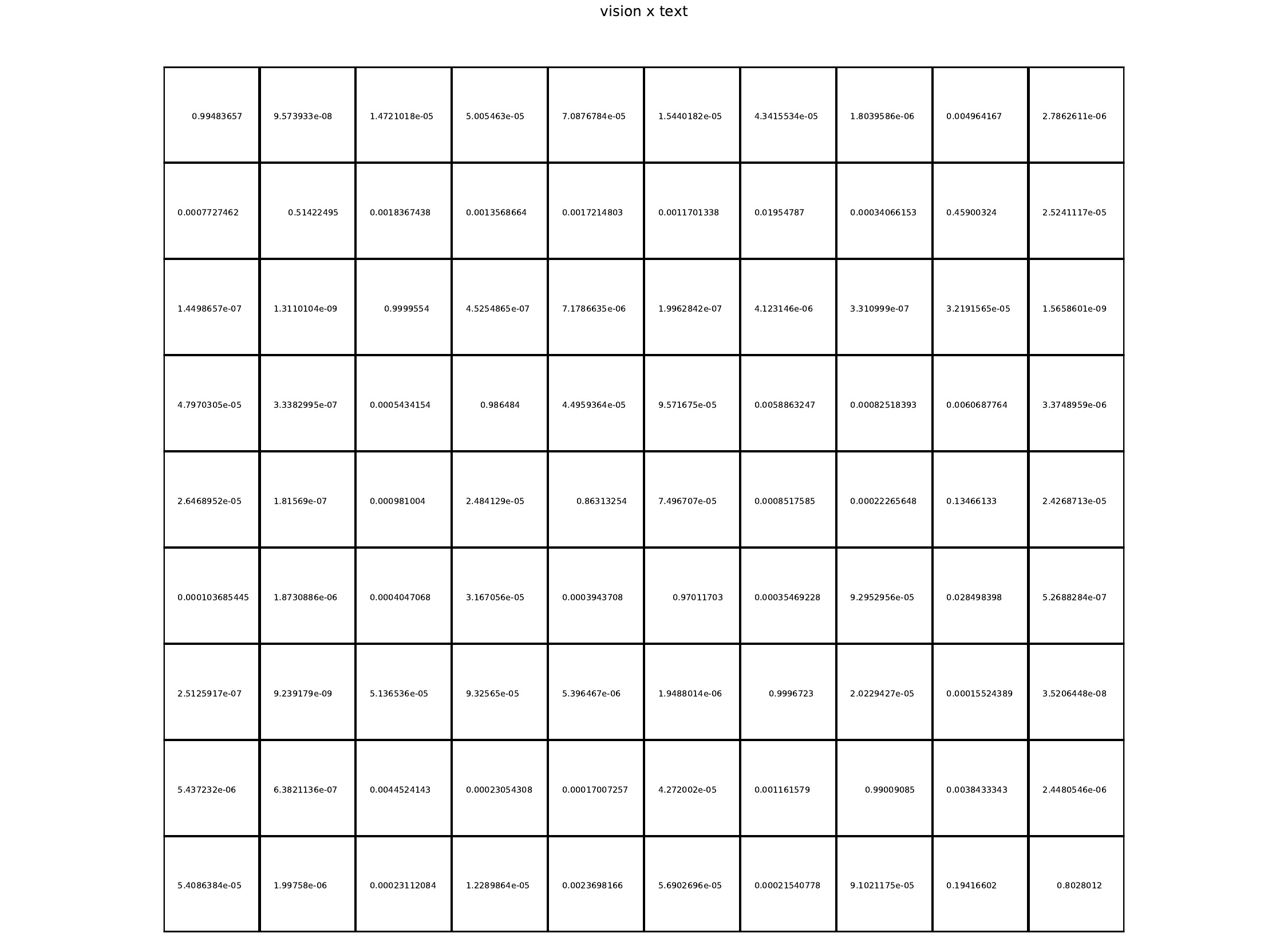}\label{fig:vt_mat_golf_clipseg}
  
  \vspace{-0.10in}
  \caption{
  Examples obtained using the proposed framework
  for different multimodal models, such as CLIPSeg. The results are given in the same format as depicted in Fig. 3 in the main paper. The example demonstrates that the method is model-agnostic.
}
  \label{fig:overall_golf_clipseg}
  \vspace{-0.20in}
\end{figure*}

\begin{figure*}[ht]
  \centering
  \includegraphics[width=1.0\textwidth]{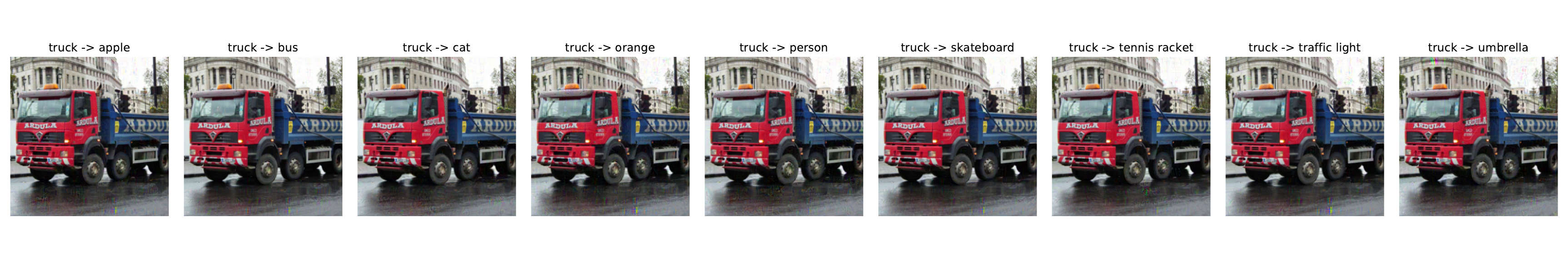}\label{fig:more_matched_mscoco}
  \includegraphics[width=1.0\textwidth]{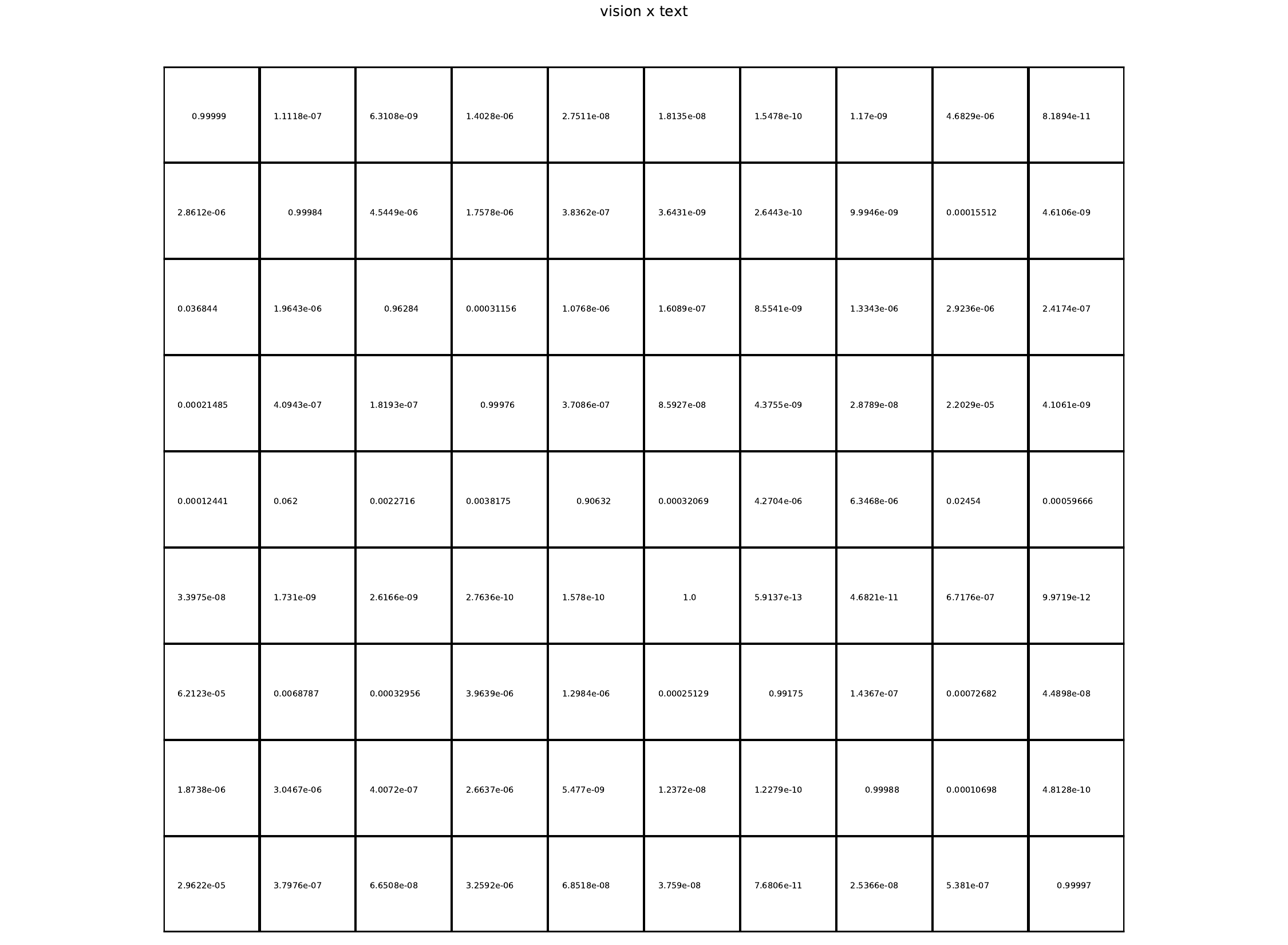}\label{fig:vt_mat_truck_mscoco}
  
  \vspace{-0.10in}
  \caption{
  Examples obtained using the proposed framework
  for the MS-COCO dataset. The results are given in the same format as depicted in Fig. 3 in the main paper. The example demonstrates that the method is dataset-agnostic.
}
  \label{fig:overall_truck_mscoco}
  \vspace{-0.20in}
\end{figure*}

\begin{figure*}[ht]
  \centering
  \includegraphics[width=1.0\textwidth]{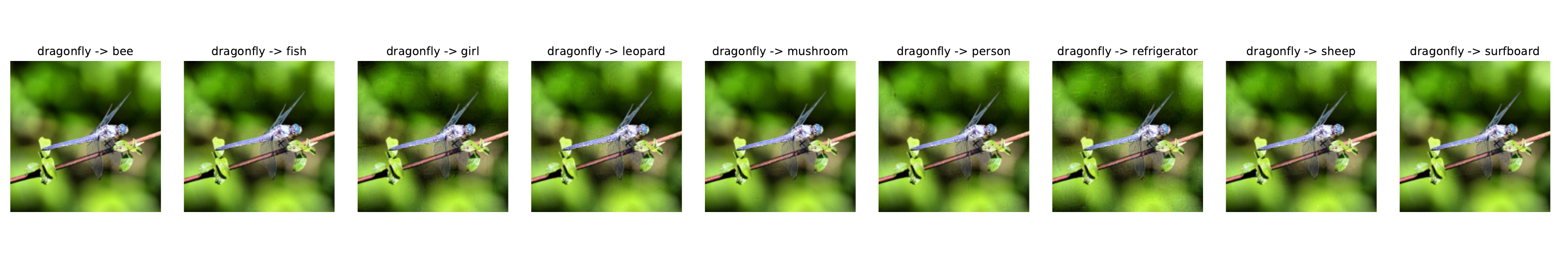}\label{fig:more_matched_goi}
  \includegraphics[width=1.0\textwidth]{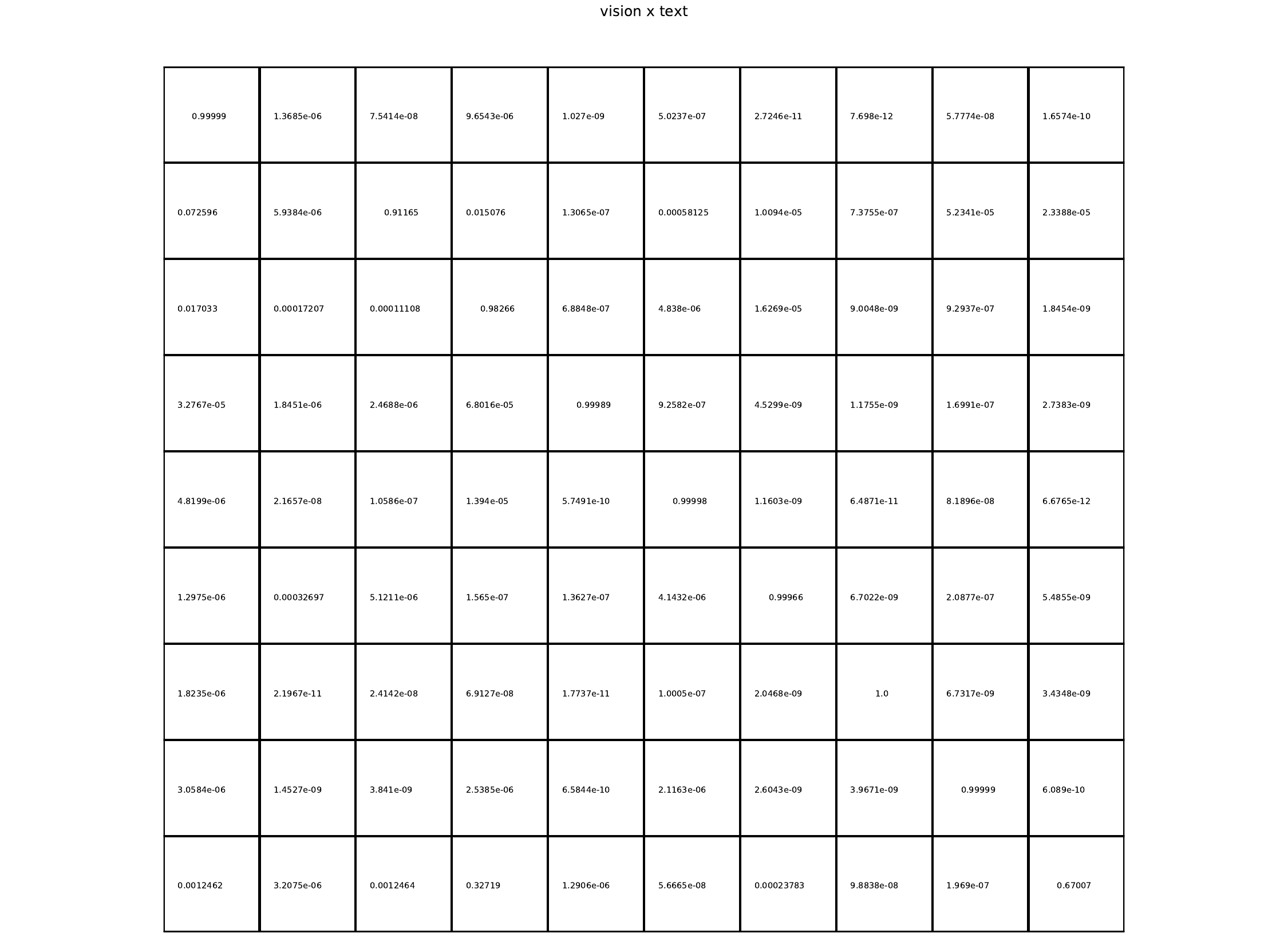}\label{fig:vt_mat_dragonfly_goi}
  
  \vspace{-0.10in}
  \caption{
  Examples obtained using the proposed framework
  for the Open Images dataset. The results are given in the same format as depicted in Fig. 3 in the main paper. The example demonstrates that the method is dataset-agnostic.
}
  \label{fig:overall_dragonfly_goi}
  \vspace{-0.20in}
\end{figure*}

Fig.~\ref{fig:conf_matrix_image_wang} to Fig.~\ref{fig:conf_matrix_open-images} show the confusion matrices for different datasets (such as Imagewang \footnotemark\footnotetext{https://github.com/fastai/imagenette}, MS-COCO, and Google Open-Images). Once again, we utilize the pretrained ImageBind model for classification. In Fig.~\ref{fig:conf_matrix_image_wang}, the representation includes 20 classes, with 10 considered as easily classified and the remaining representing various dog breeds. The overall zero-shot accuracy stands at $75.37\%$. Notably, the distinct impact of different dog breed classes on the overall zero-shot performance is visually evident. 
For Fig.~\ref{fig:conf_matrix_MS_coco} and Fig.~\ref{fig:conf_matrix_open-images}, we randomly select classes from both datasets. The overall accuracy stands at $62.9\%$ for MS-COCO and $68.88\%$ for Open-Images.
In general, these multimodal models work well when the classes are very different. However, their effectiveness becomes more limited in scenarios where subtle differences exist between the classes.

We extend our systematic evaluation to all these datasets, repeating the process similarly to what is performed for Imagenette in the main paper. As depicted in the main paper results, we successfully match all of them to the wrong class with visually imperceptible changes.
Fig. \ref{fig:overall_truck_mscoco} and Fig. \ref{fig:overall_dragonfly_goi} show two examples with MS-COCO and Open Images respectively, using the pretrained multimodal ImageBind model. Fig. \ref{fig:overall_golf_clipseg} shows another example, that uses a different multimodal pretrained model, known as CLIPSeg. The results are given in the same format as depicted in Fig. 3 in the main paper. While the images are visually indistinguishable, their
representations given by the ImageBind or CLIPSeg model are very different. The classification probabilities given by the {\em unmodified} ImageBind or CLIPSeg model are shown in the corresponding vision $\times$ text matrix. The probabilities indicate that these images are classified with very high confidence into the matched class rather than the actual class to which the image belongs. Consequently, the classification accuracy of the same multimodal model is 0\%.

In Table \ref{psnr_ssim_res}, we present the average PSNR (peak signal-to-noise ratio) and SSIM (structural similarity index measure) values between the original and manipulated (i.e., embedding-aligned) images across all considered datasets. These metrics serve as effective measures of image similarity, indicating that the image quality does not significantly degrade with minimal distortion.

In Fig.~\ref{fig:diffnoise}, the actual difference images for three typical examples of visually indistinguishable pairs are shown. We observe that the differences are indeed very small, aligning consistently with the high PSNR and SSIM Index values presented in Table 1.


\vfill

\end{document}